\pdfoutput=1

\documentclass[11pt]{article}

\usepackage[final]{acl}

\usepackage{times}
\usepackage{latexsym}

\usepackage[T1]{fontenc}

\usepackage[utf8]{inputenc}

\usepackage{microtype}

\usepackage{inconsolata}

\usepackage{graphicx}

\usepackage[cjk]{kotex}
\usepackage[most]{tcolorbox}
\usepackage[normalem]{ulem}
\usepackage{amsmath}
\usepackage{amssymb}
\usepackage{array}
\usepackage{arydshln}
\usepackage{bm}
\usepackage{booktabs}
\usepackage{CJKutf8}
\usepackage{cleveref}
\usepackage{colortbl}
\usepackage{diagbox}
\usepackage{footmisc}
\usepackage{keystroke}
\usepackage{lipsum}
\usepackage{makecell}
\usepackage{mdframed}
\usepackage{menukeys}
\usepackage{multirow}
\usepackage{pifont}
\usepackage{slashbox}
\usepackage{soul}
\usepackage{subcaption}
\usepackage{tabularx}
\usepackage{tikz}
\usepackage{upquote}
\usepackage{xspace}
\usepackage{booktabs}
\usepackage{graphicx}
\usepackage{afterpage}
\usepackage{float}
\usepackage{placeins}
\PassOptionsToPackage{table}{xcolor}
\usepackage{xcolor}

\graphicspath{ {./images/} }
\mdfsetup{
  linecolor=white,
  backgroundcolor=gray!20,
}
\newcolumntype{?}{!{\vrule width 1.2pt}}
\useunder{\uline}{\ul}{}
\crefformat{section}{\S#2#1#3}
\crefformat{subsection}{\S#2#1#3}
\crefformat{subsubsection}{\S#2#1#3}

\newcommand{\cmark}{\ding{51}}
\newcommand{\xmark}{\ding{55}}
\newcommand{\tool}{\textbf{\textsc{LLM-C3Mod}}\xspace}

\title{\tool: A Human-LLM Collaborative System for Cross-Cultural Hate Speech Moderation}

\author{
  Junyeong Park$^{\diamond,*}$,
  Seogyeong Jeong$^{\diamond,*}$,
  Seyoung Song$^{\diamond,}$\Thanks{ Equal contribution.},
  Yohan Lee$^{\diamond,\dagger}$,
  Alice Oh$^\diamond$
  \\
  \ \\
  $^\diamond$KAIST, $^\dagger$ETRI\\
  \texttt{\{\href{mailto:jjjunyeong9986@kaist.ac.kr}{\color{black}{jjjunyeong9986}}, \href{mailto:sg.jeong28@kaist.ac.kr}{\color{black}{sg.jeong28}}, \href{mailto:seyoung.song@kaist.ac.kr}{\color{black}{seyoung.song}}\}@kaist.ac.kr},\\
  \texttt{carep@etri.re.kr, alice.oh@kaist.edu}
}

\begin{document}
\maketitle
\begin{abstract}
  \textit{\textbf{Warning}: This paper contains content that may be offensive or upsetting}

Content moderation is a global challenge, yet major tech platforms prioritize high-resource languages, leaving low-resource languages with scarce native moderators. Since effective moderation depends on understanding contextual cues, this imbalance increases the risk of improper moderation due to non-native moderators' limited cultural understanding.
Through a user study, we identify that non-native moderators struggle with interpreting \textit{culturally-specific knowledge}, \textit{sentiment}, and \textit{internet culture} in the hate speech moderation.
To assist them, we present \tool, a human-LLM collaborative pipeline with three steps: (1) RAG-enhanced cultural context annotations; (2) initial LLM-based moderation; and (3) targeted human moderation for cases lacking LLM consensus.
Evaluated on a Korean hate speech dataset with Indonesian and German participants, our system achieves 78\% accuracy (surpassing GPT-4o's 71\% baseline), while reducing human workload by 83.6\%.
Notably, human moderators excel at nuanced contents where LLMs struggle.
Our findings suggest that non-native moderators, when properly supported by LLMs, can effectively contribute to cross-cultural hate speech moderation.

\end{abstract}

\section{Introduction}
\label{sec:introduction}

\begin{figure}[t]
  \centering
  \includegraphics[width=\linewidth]{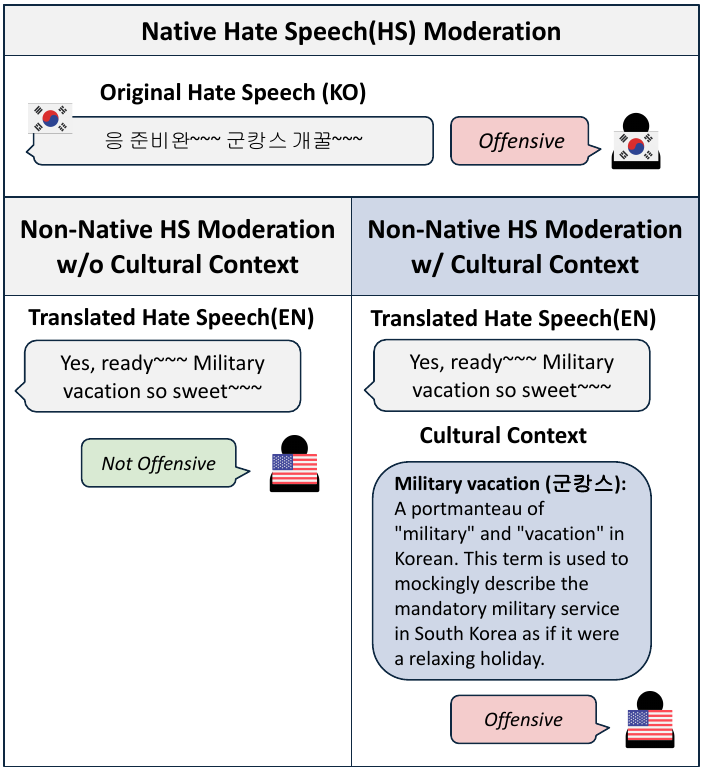}
  \caption{An example of a non-native hate speech moderator performing hate speech detection with and without cultural context.}
  \label{fig:motivation}
\end{figure}

Content moderation has evolved into a global challenge, yet major tech platforms concentrate their resources primarily on high-resource languages~\cite{witness2023big}.
Meta allocates 87\% of its misinformation budget to English content despite only 9\% of users being English speakers, exemplifying a systemic bias in content moderation~\cite{milmo2021facebook}.
This imbalance has led to increased hate speech and misinformation in non-English contexts, alongside risks of improper content moderation due to insufficient cultural understanding~\cite{nigatu2024searched, elswah2024investigating}.

Given the scarcity of native moderators for many languages, we argue that exploring methods for non-native hate speech moderation is crucial.
As exemplified in Figure~\ref{fig:motivation}, non-native moderators cannot simply rely on machine translation, as hate speech moderation task requires deeper cultural and political context to make an informed decision~\cite{chan2024hate, lee-etal-2024-exploring-cross}.
Recent research has explored using Large Language Models (LLMs) for content moderation~\cite{kolla2024llm, jha-etal-2024-memeguard} and hate speech detection~\cite{roy-etal-2023-probing, zhang-etal-2024-dont-go}, but primarily focuses on single-language scenarios, leaving cross-cultural challenges largely unexplored~\cite{pawar2024survey, hee-etal-2024-recent}.

We present \tool, a system that leverages retrieval-augmented generation (RAG)-enhanced LLMs~\cite{lewis2020retrieval} to assist non-native moderators through three key components: (1) cultural context annotation, (2) initial LLM-based moderation, and (3) targeted human moderation.
Our system leverages web search results to generate reliable cultural annotations, helping non-native moderators better understand culturally specific expressions and nuances.
Also, through LLM-based initial screening, we maintain efficient workload distribution between automated and human moderation.

We evaluate \tool on KOLD~\cite{jeong-etal-2022-kold}, a Korean hate speech dataset, with non-native participants from Indonesia and Germany.
Our system achieves 78\% accuracy (surpassing the 71\% GPT-4o baseline) while reducing human workload by 83.6\%.
Notably, providing cultural context annotations improves non-native moderator accuracy from 22\% to 61\%.
We found that human moderators particularly excel at nuanced tasks where LLMs struggle, such as interpreting internet culture, including memes and their cultural references.

Our main contributions are as follows:

\begin{itemize}
  \item We empirically identify key challenges faced by non-native moderators in cross-cultural hate speech moderation through user study.
  \item We develop a RAG-enhanced cultural annotation system that demonstrably improves hate speech moderation accuracy for both humans and LLMs.
  \item We propose \tool, an effective human-LLM collaboration pipeline that strategically integrates machine efficiency with human judgment.
\end{itemize}

Our findings demonstrate that non-native moderators, when properly supported by LLMs, can contribute effectively to cross-cultural hate speech moderation, addressing critical needs in global online safety.

\section{Related Work}
\label{sec:relwork}
\subsection{Hate Speech Moderation: Cultural Considerations}

Hate speech moderation is a type of content moderation that involves various tasks, including detecting~\cite{park-fung-2017-one, vidgen-etal-2021-learning}, explaining~\cite{sap-etal-2020-social, elsherief-etal-2021-latent, Mathew_Saha_Yimam_Biemann_Goyal_Mukherjee_2021}, and countering~\cite{10.1145/3534678.3539161, chung-etal-2019-conan} hate speech on online platforms.
One of the challenges in this domain lies in understanding diverse cultural and contextual cues that differ across countries and regions~\cite{hee-etal-2024-recent}.

To address this challenge, recent works have introduced hate speech datasets that incorporate various linguistic and cultural factors~\cite{lee-etal-2023-hate, jeong-etal-2022-kold, jin-etal-2024-kobbq, lee-etal-2024-exploring-cross, arango-monnar-etal-2022-resources, deng-etal-2022-cold, demus-etal-2022-comprehensive, maronikolakis-etal-2022-listening, ye2024federatedapproachfewshothate, muhammad2025afrihatemultilingualcollectionhate}.
Another recent works have proposed culturally-specific hate speech moderation methods~\cite{li2024cultureparkboostingcrossculturalunderstanding, ye2024federatedapproachfewshothate}. Furthermore, \citet{masud-etal-2024-hate} explore the potential of utilizing LLMs as hate speech annotators representing specific cultural or geographical groups.
However, these approaches largely focus on moderation within specific mono-cultural contexts. This leaves a gap in addressing the complexities of cross-cultural hate speech moderation where human moderators are required to handle content from unfamiliar cultural or linguistic contexts. In this work, we examine the difficulties of non-native annotators and their potential in cross-cultural hate speech moderation.

\subsection{Hate Speech Moderation: Human-LLM Collaboration}

Recent works have investigated LLM-assisted content moderation~\cite{10.1145/3613905.3650828, Kumar_AbuHashem_Durumeric_2024} and hate speech moderation~\cite{10646809, kang-qian-2024-implanting, 10.24963/ijcai.2023/694, yang-etal-2023-hare, podolak-etal-2024-llm}.
However, for tasks that are heavily context-dependent, such as content moderation, human moderators are known to outperform automated systems by making more nuanced decisions that consider contextual subtleties ~\cite{10.1145/3290605.3300760, doi:10.1177/2053951719897945}.

Thus, to utilize both human intelligence and machine moderator's scalability and efficiency, there is growing exploration of human-machine collaboration for hate speech moderation~\cite{10.1145/3338243, thomas2024supportinghumanratersdetection, ding2024counterquillinvestigatingpotentialhumanai, breazu2024largelanguagemodelsthematic}. Yet, it remains unclear how LLMs can be effectively leveraged in cross-cultural hate speech moderation scenarios. In this work, we utilize LLMs as cultural context annotators and hate speech moderator agents, proposing a human-LLM collaboration cross-cultural hate speech moderation pipeline.

\section{User Study: Understanding Non-Native Moderators' Challenges}
\label{sec:challenges}
In this section, we explore the challenges non-native moderators face when relying solely on basic machine translation for hate speech detection. A user study was conducted with non-Korean moderators on KOLD~\cite{jeong-etal-2022-kold}, a Korean hate speech detection dataset.

\subsection{Method}
\label{sec:user_study_method}

\paragraph{Dataset}
KOLD~\cite{jeong-etal-2022-kold} consists of comments and titles from Naver News and YouTube, annotated by native Korean speakers for offensiveness.
From this dataset, we manually curated 100 culturally specific samples and categorized them into 8 themes including political, religious, historical topic.
For each theme, one offensive and one non-offensive sample were selected, resulting in 16 samples for the user study.
The samples were translated into English using GPT-4o~\cite{hurst2024gpt}, creating 16 English comment-title pairs for evaluation.

\paragraph{Experimental Design}
In this user study, two non-Korean graduate students participated as annotatiors. One student is from Indonesia and the other student is from Germany. Neither had prior exposure to the KOLD dataset.

The participants' task was to annotate the offensiveness of the provided comments following adapted guidelines based on the KOLD dataset annotation framework.
These guidelines, as in the original KOLD guideline, included identifying and marking specific spans of text considered offensive within the comments.
Aside from the usual ``Offensive'' and ``Non-offensive'' options, we introduced an additional ``I don't know'' option. Specifically, when the participant is uncertain about a comment's offensiveness, they were instructed to select ``I don't know'' and indicate what additional information would help them make a decision.
Also, the participants were permitted to use an English dictionary for clarifying word meanings but were strictly prohibited from using search engines or LLMs during the annotation process.

\subsection{Results}
The participants struggled with the task, answering incorrectly or selecting ``I don't know'' for nearly half of the samples, achieving an overall accuracy of 56.25\%. Participant 1 answered correctly for 9 samples, incorrectly for 2, and chose ``I don't know'' for 5 samples. Similarly, participant 2 answered correctly for 9 samples, incorrectly for 4, and chose ``I don't know'' for 3 samples.

\begin{figure*}[t]
  \centering
  \includegraphics[width=\textwidth]{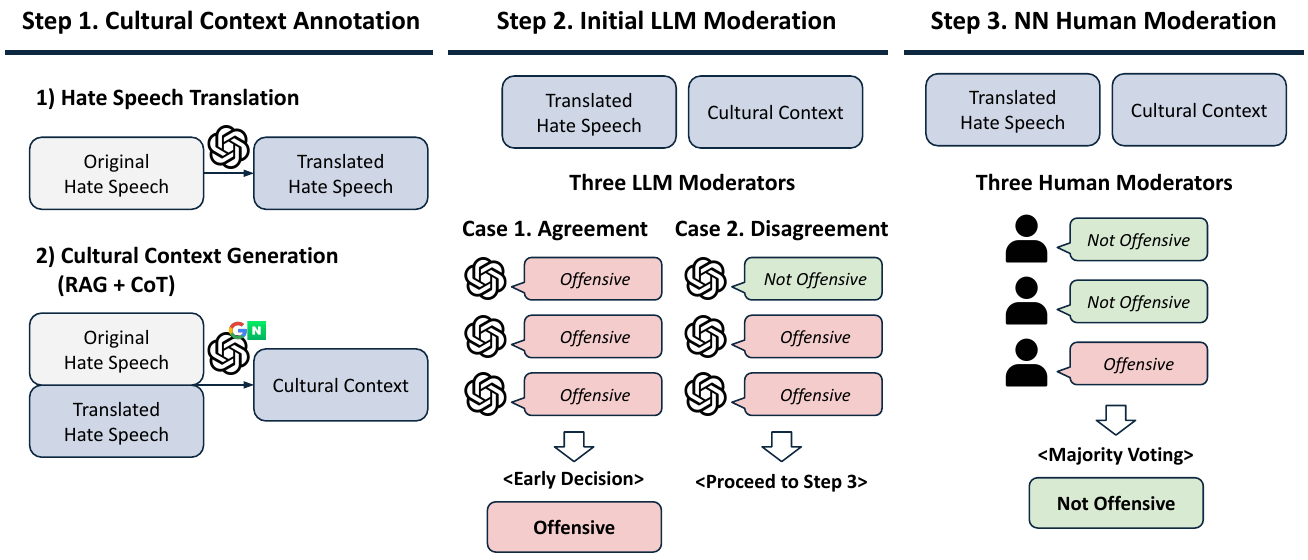}
\caption{Overview of \tool. The pipeline consists of three steps: 1) generating cultural context annotations, 2) initial moderation using LLM moderators, and 3) final moderation by non-native human moderators. Further details are provided in Section~\ref{sec:pipeline}.}
\label{fig:collaboration_pipeline}
\end{figure*}

\subsection{Findings}
\label{sec:user_study_findings}

The user study revealed three key challenges faced by non-native moderators: difficulties in understanding \textit{culturally-specific knowledge}, \textit{sentiment} and \textit{internet culture}.

\paragraph{Cultural Knowledge}
Participants struggled with unfamiliar Korean-specific named entities such as ``Northwest Youth League (서북청년단),''.
For instance, in the comment ``If it were our country, it would be like the Northwest Youth League ruling the nation (우리나라로 치면 서북청년단이 나라를 지배하는 꼴)'', both participants selected ``I don't know'' and indicated that they need more information about the named entity ``Northwest Youth League''.

\paragraph{Cultural Sentiment}
Another challenge arose from the cultural sentiment disparities.
For example, participants marked the comment ``root out pro-Japanese collaborators (친일파를 뿌리 뽑다)'' as ``offensive'' due to the phrase ``root out''.
However, in the Korean cultural context, ``pro-Japanese collaborators'' refers to individuals who cooperated with Japanese imperial policies during the colonial era, a group widely criticized and condemned in Korea.
Thus, the comment is considered non-offensive within its cultural context. However, these participants marked it as offensive because they did not share the sentiments and cultural sensitivity of Koreans.

\paragraph{Internet Culture}
The participants also encountered difficulties with understanding Korean internet memes, slang, and humor such as the comment ``The reason why Gag Concert has no choice but to fail...(개콘이 망할 수 밖에 없는 이유...)''. Gag Concert, a popular Korean comedy show, is often referenced in internet memes to describe absurd real-life situations, especially in serious contexts like politics or religion. The meme suggests that these real events are so ironic and comedic that they outshine scripted humor, causing the comedy show to seem less relevant.
Both participants marked ``I don't know'' due to a lack of context to understand the reference.

These findings emphasize the need to provide cultural context for non-native moderators in hate speech detection tasks, especially to assist them in understanding \textit{culturally-specific} \textit{knowledge}, \textit{sentiment}, and \textit{internet culture}. Hate speech examples for each category are provided in Appendix~\ref{appendix:AA_cultura_context_sample}.

\section{\tool: A Human-LLM Collaborative Hate Speech Moderation Pipeline}
\label{sec:pipeline}

\begin{table*}[]
  \resizebox{\textwidth}{!}{%
    \begin{tabular}{lllrrr}
      \toprule
      &
      &
      &
      \multicolumn{1}{c}{} &
      \multicolumn{1}{c}{} &
      \multicolumn{1}{c}{} \\
      &
      &
      &
      \multicolumn{1}{c}{\multirow{-2}{*}{
          \begin{tabular}[c]{@{}c@{}}Number of\\ Samples
      \end{tabular}}} &
      \multicolumn{1}{c}{\multirow{-2}{*}{
          \begin{tabular}[c]{@{}c@{}}Baseline \\ (GPT-4o)
      \end{tabular}}} &
      \multicolumn{1}{c}{\multirow{-2}{*}{
          \begin{tabular}[c]{@{}c@{}}Our Pipeline \\ (GPT-4o \& Human)
      \end{tabular}}} \\ \midrule
      &
      \multicolumn{2}{l}{\textbf{All Samples}} &
      \cellcolor[HTML]{FFFC9E}{\color[HTML]{000000} 171} &
      \cellcolor[HTML]{FFFC9E}{\color[HTML]{000000} 0.71} &
      \cellcolor[HTML]{FFFC9E}{\color[HTML]{000000} \textbf{0.78}} \\
      &
      \multicolumn{2}{l}{Decision at Step 2: LLM Moderators} &
      \cellcolor[HTML]{FFFC9E}{\color[HTML]{000000} 143} &
      \cellcolor[HTML]{FFFC9E}{\color[HTML]{000000} 0.72} &
      \cellcolor[HTML]{FFFC9E}{\color[HTML]{000000} \textbf{0.78}} \\
      \multirow{-3}{*}{\textbf{Total}} &
      \multicolumn{2}{l}{Decision at Step 3: Human Moderators} &
      \cellcolor[HTML]{FFFC9E}{\color[HTML]{000000} 28} &
      \cellcolor[HTML]{FFFC9E}{\color[HTML]{000000} 0.67} &
      \cellcolor[HTML]{FFFC9E}{\color[HTML]{000000} \textbf{0.75}} \\ \hline
      &
      \multicolumn{2}{l}{\textbf{All Samples}} &
      61 &
      \textbf{0.78} &
      0.75 \\
      &
      \multicolumn{2}{l}{Decision at Step 2: LLM Moderators} &
      54 &
      \textbf{0.76} &
      \textbf{0.76} \\
      \multirow{-3}{*}{
        \begin{tabular}[c]{@{}l@{}}\textit{Cultural} \textit{Knowledge}
      \end{tabular}} &
      \multicolumn{2}{l}{Decision at Step 3: Human Moderators} &
      7 &
      \textbf{0.91} &
      0.71 \\ \hline
      &
      \multicolumn{2}{l}{\textbf{All Samples}} &
      51 &
      0.69 &
      \textbf{0.78} \\
      &
      \multicolumn{2}{l}{Decision at Step 2: LLM Moderators} &
      41 &
      0.76 &
      \textbf{0.78} \\
      \multirow{-3}{*}{
        \begin{tabular}[c]{@{}l@{}}\textit{Cultural} \textit{Sentiment}
      \end{tabular}} &
      \multicolumn{2}{l}{Decision at Step 3: Human Moderators} &
      10 &
      0.43 &
      \textbf{0.80} \\ \hline
      &
      \multicolumn{2}{l}{\textbf{All Samples}} &
      59 &
      0.6 &
      \textbf{0.80} \\
      &
      \multicolumn{2}{l}{Decision at Step 2: LLM Moderators} &
      48 &
      0.65 &
      \textbf{0.81} \\
      \multirow{-3}{*}{
        \begin{tabular}[c]{@{}l@{}}\textit{Internet} \textit{Culture}
      \end{tabular}} &
      \multicolumn{2}{l}{Decision at Step 3: Human Moderators} &
      11 &
      \textbf{0.73} &
      \textbf{0.73} \\ \bottomrule
    \end{tabular}%
  }
\caption{Comparison of \tool(GPT-4o \& Non-native Human) and a GPT-4o baseline(avg. of three runs) on 171 KOLD dataset samples. The samples are categorized based on the required type of cultural understanding: 1) cultural knowledge (N= 61), 2) cultural
sentiment(N = 51), and 3) internet culture(N = 59). Using \tool, samples are divided into two groups:
those resolved in Step 2 with agreement among LLM moderators and those requiring further
review by human moderation in Step 3. \tool significantly improves performance in Step 3,
increasing overall accuracy from 0.71 to 0.78. KOLD samples for each category, along with cultural context annotations, are provided in Appendix~\ref{appendix:AA_cultura_context_sample}.}
\label{tab:main-table}
\end{table*}

In this section, we suggest how LLMs can assist non-native moderators in understanding and moderating cross-cultural hate speech.

Based on our findings in Section~\ref{sec:challenges}, we propose \tool, a human-LLM collaborative hate speech moderation pipeine that includes 1) automatically generating cultural context 2) initial moderation with LLM moderators and 3) moderation with non-native human moderators. The process is described in Figure~\ref{fig:collaboration_pipeline}.

\paragraph{Step 1: Automatic Cultural Context Generation}
To assist hate speech moderation, we automatically generate cultural context of each title-comment pairs with GPT-4o~\cite{hurst2024gpt}. Notably, reliable cultural context annotations should not contain misinformation and should be able to handle up-to-date information, considering the real-time nature of content moderation. However, LLMs have limitations as they cannot process data beyond their training time and exhibit inherent hallucination \cite{xu2024hallucination}.

To mitigate these problems, we employ RAG~\cite{lewis2020retrieval} and CoT~\cite{NEURIPS2022_9d560961} frameworks. Specifically, we use following steps to generate cultural context annotation: (1) detect text span in the titles and comments related to following three aspects---\textit{culturally-specific knowledge}, \textit{sentiment}, and \textit{internet culture}; (2) search for related articles or documents in the internet(RAG); (3) annotate objective cultural context based on the retrieved information. The samples of generated cultural context are shown in Appendix~\ref{appendix:AA_cultura_context_sample}. Furthermore, the prompts used in this process and their corresponding responses are detailed in Appendix \ref{appendix:prompt_annotation} and \ref{appendix:D_annotation_generation_example}, respectively.

Since our goal is to provide additional information that can assist non-native moderators in making accurate decision, we strictly limit our annotation to `objective contexts'. In this stage, we do not task LLMs with determining whether a comment is offensive.

\paragraph{Step 2: Initial LLM Moderation}
To ensure scalability of the pipeline, we employ LLM agents for initial hate speech detection. Using the cultural context annotations generated in Step 1, three LLM moderators classify each comment as either offensive or non-offensive. The outcomes fall into one of two scenarios: (1) all three LLM moderators agree, or (2) one LLM moderator disagrees with the other two. In the first case, the pipeline concludes with the unanimous decision of the LLM moderators. In the second case, the pipeline moves to the next step for further review. In this study, we utilized three GPT-4o~\cite{hurst2024gpt} agents as LLM moderators.

\paragraph{Step 3: Non-native Human Moderation}
Samples flagged due to LLM disagreement are passed to non-native human moderators, as such samples are implicitly more challenging. Human moderators are provided with the same cultural context annotations, titles and comments. The final decision is determined by majority voting among three non-native human moderators.

\section{Experiments}
\label{sec:experiments}

\subsection{Cultural Context Annotation}

We conduct an A/B test to evaluate the effentiveness of cultural context annotations using a small set of 12 manually selected samples from the KOLD dataset. The samples include seven offensive and five non-offensive comments, four from each category---\textit{culturally-specific knowledge}, \textit{sentiment}, and \textit{internet culture}.
For human moderators, we recruited three non-Korean participants. Initially, they performed hate speech detection without the cultural context annotations, following the procedure described in Section \ref{sec:user_study_method}. Then, they repeated the task on the same set of samples with the cultural context annotations provided. We conducted the same task using three GPT-4o moderators.

\begin{table}[t]
  \centering
  \begin{tabular}{@{}ccc@{}}
    \toprule
    & \multicolumn{2}{c}{Cultural Context Annotation} \\
    & \xmark             & \cmark             \\ \midrule
    Human Moderators  & 0.22               & \textbf{0.61}               \\
    GPT-4o Moderators & 0.67               & \textbf{0.92}               \\ \bottomrule
  \end{tabular}
  \caption{Performance of humans and LLMs in hate speech detection with and without cultural context annotations on 12 KOLD samples. The performance is measured as the average of three non-native human moderators and three GPT-4o moderators.}
  \label{tab:annotation}
\end{table}

\begin{table}[]
  \resizebox{\columnwidth}{!}{%
    \begin{tabular}{@{}llll@{}}
      \toprule
      & Avg. Acc. & Agree. Ratio & Agree. Acc. \\ \midrule
      GPT-4o                & \textbf{0.74} & \textbf{0.84}  & \textbf{0.75}    \\
      Claude-3-haiku        & 0.71     & \textbf{0.84}            & 0.73              \\
      Gemini-1.5            & 0.73     & 0.82            & 0.74              \\
      Mixed & 0.72     & 0.78                & 0.72                  \\ \bottomrule
    \end{tabular}%
  }
  \caption{Comparison of LLM Moderator Groups – Each group consists of three GPT-4o, Claude-3-Haiku, Gemini-1.5, or a mix of these models. Avg. Acc. represents the average hate speech detection accuracy. Agree. Ratio indicates the proportion of samples with unanimous agreement among all models in a group. Agree. Acc. measures accuracy on those unanimously agreed samples.}
  \label{tab:llms-accuracy}
\end{table}

Table~\ref{tab:annotation} shows that the generated cultural context annotations help improve the performance of both humans and LLMs in hate speech detection task.
In particular, LLMs demonstrate high accuracy when the annotations are provided, showing promises.

\subsection{LLM Moderators}
We compare moderation capabilities of various LLMs to determine the most suitable LLM to serve as the moderator in our pipeline. For this section and the evaluation of pipeline, we manually select 171 samples from the KOLD dataset. Specifically, 50 samples were categorized as \textit{cultural knowledge}, 62 as \textit{cultural sentiment}, and 60 as \textit{internet culture}.

Aligned with our proposed pipeline, we evaluate three LLMs as a group and compare their agreement ratios and accuracy on unanimously agreed answers. The comparison includes a GPT-4o group, a Claude-3-haiku group, a Gemini-1.5 group, and a mixed group consisting of one GPT-4o, one Claude-3-haiku, and one Gemini-1.5.

In Table~\ref{tab:llms-accuracy}, the results show that GPT-4o group achieves the highest average accuracy. While Claude-3-haiku group demonstrates the highest agreement ratio, it falls short in accuracy, making it the least suitable option for our pipeline. GPT-4o achieves the best accuracy on samples where unanimous agreement is reached. Although GPT-4o group reaches unanimous agreement on fewer samples, the accuracy of its agreed-upon samples is high, the high accuracy of these agreed-upon samples makes it a reliable choice for our pipeline. Based on these findings, we use three GPT-4o agents as the LLM moderators in our pipeline.

\subsection{\tool Pipeline}
The goal of this pipeline is to accurately and effectively conduct hate speech moderation. Based on prior findings, GPT-4o is employed as both the cultural annotation generator and the LLM moderator. For non-native human moderators, we recruited three graduate students: two from Indonesia and one from Germany. We use the same 171 KOLD samples from the LLM moderator evaluation experiment.

Table~\ref{tab:main-table} compares the performance of our pipeline with a GPT-4o baseline (avg. of three runs). Our pipeline achieved 78\% accuracy, exceeding the GPT-4o baseline accuracy of 71\%. Furthermore, only 28 out of 171 samples failed to achieve unanimous agreement among the LLM moderators, reducing the workload for human moderators by 83.6%

In Step 2, of the 143 samples that reached unanimous agreement, the LLM moderators made correct decisions on 112 samples, achieving 78\% accuracy. In Step 3, majority voting among non-native human moderators achieved 75\% accuracy, significantly surpassing the baseline GPT-4o's accuracy of 43\%. These results demonstrate that our pipeline effectively improves the overall performance of hate speech moderation by identifying more challenging samples and delegating them to human moderators for review.

The performance of our pipeline showed no significant differences across the three categories (Table~\ref{tab:main-table}).
However, there were several interesting features when our pipeline (human-LLM collaboration) is compared to the baselines.
First, in the \textit{cultural knowledge} category, where extracting factual data is more critical than understanding nuances, the performance decreased after applying our pipeline.
However, in the \textit{cultural sentiment} category and \textit{internet culture} category, where understanding nuances takes precedence, the performance significantly improve through our pipeline.
The accuracy comparison within the actual pipeline, specifically between the three LLM moderators and the non-native human moderator in step 3 (majority voting) can be seen in Table \ref{tab:category_eachLLM_stage3}. For \textit{cultural knowledge}, the Non-native human moderator accuracy shows significant fluctuation, sometimes higher and sometimes lower. However, for other categories, the accuracy generally tends to improve. In the case of \textit{internet culture} category, while the final LLM moderator accuracy is slightly higher than the human moderator accuracy, this difference is only by one sample among 11 samples. When considering the overall performance across the three LLM moderators, the NN-human moderator case generally shows an upward trend in \textit{internet culture} category.

These observations suggest that in content moderation tasks, there are aspects where humans still outperform LLMs by a substantial margin especially when understanding context and nuance is critical.

\begin{table}[t]
  \resizebox{\columnwidth}{!}{
    \centering
    \begin{tabular}{lrrrr}
      \toprule
      & \multicolumn{3}{c}{
        \begin{tabular}[c]{@{}c@{}}LLM Moderator\\ (GPT-4o)
      \end{tabular}} & \multicolumn{1}{c}{\multirow{2}{*}{
          \begin{tabular}[c]{@{}c@{}}NN Human\\ Moderator
      \end{tabular}}} \\
      & \multicolumn{1}{c}{1}      & \multicolumn{1}{c}{2}      & \multicolumn{1}{c}{3}      & \multicolumn{1}{c}{}                                                                              \\ \hline
      \textbf{Total}                & \colorbox{blue!40}{0.43}                        & \colorbox{blue!40}{0.57}                      & \colorbox{blue!40}{0.61}                       & 0.75                                                                                     \\
      \textit{Cultural Knowledge}         & \colorbox{red!40}{0.86}                      & 0.71 & \colorbox{blue!40}{0.43}                       & 0.71                                                                                              \\
      \textit{Cultural Sentiment} & \colorbox{blue!40}{0.30}               & 0.80 & \colorbox{blue!40}{0.50} & 0.80                                                                                      \\
      \textit{Internet Culture}          & \colorbox{blue!40}{0.27} & \colorbox{blue!40}{0.27}                       & \colorbox{red!40}{0.82}              & 0.72       \\ \hline
    \end{tabular}
  }
  \caption{Accuracy comparison in the Step3 in our pipeline: 3 LLM moderators(GPT-4o)' accuracy and Majority voting accuracy between 3 non-native human moderators. The comparision was done on 28 samples, and on each category; named entity (N=7), cultural sensitivity (N=10), and local memes (N=11). Cases where the LLM Moderator accuracy is lower than the NN-Human Moderator's Majority Voting accuracy are highlighted in \colorbox{blue!40}{blue}, while cases where it is higher are highlighted in \colorbox{red!40}{red}.}
  \label{tab:category_eachLLM_stage3}
\end{table}

\section{Discussion}
\label{sec:discussion}
\subsection{Native vs. Non-native Moderator Performance}

\begin{table}[]
  \resizebox{\columnwidth}{!}{%
    \begin{tabular}{@{}lccc|cc@{}}
      \toprule
      & \multicolumn{3}{c|}{Non-Native Moderators} & \multicolumn{2}{c}{Avg.} \\
      & 1 & 2 & 3 & Non-Natives &Natives \\ \midrule
      Acc. & 0.68 & 0.82 & 0.68 & \underline{0.73} & \underline{0.89} \\ \bottomrule
    \end{tabular}%
  }
  \caption{Comparison of hate speech detection accuracy between individual non-native moderators and native moderators. For non-native moderators, accuracy is calculated based on 28 samples from the pipeline validation experiment (Step 3). For native moderators, the average accuracy is calculated across 1,749 annotators who annotated more than 9 samples, using the entire KOLD dataset.}
  \label{tab:pipeline_participant_stats}
\end{table}

In this discussion section, we aim to compare the performance of non-native moderators to native moderators. We conduct a statistical analysis of Korean (native) annotators in the KOLD dataset and non-native participants in our experiment.

The hate speech detection accuracy of each individual annotator in the KOLD dataset was measured as follows. Each sample in the KOLD dataset includes the judgment results of three Korean annotators, along with their respective annotator IDs. Using this information, we identified all annotator IDs who annotated more than 9 samples from the KOLD dataset annotations. Then, we calculated the accuracy of each annotator by measuring how often their annotations matched the golden answers. The results are visualized in Figure \ref{fig:kold_korean_stat}.

As a result, we found that a total of 3,124 annotators contributed to annotating 40,429 samples in the KOLD dataset. on average, each annotator annotated 38.8 samples, with a median of 12 samples per annotator. Among them, 1,749 annotators annotated more than 9 samples. Within these filtered annotators, the mean accuracy was 0.89 (standard deviation: 0.074), and the median accuracy was 0.91. Note that the average accuracy cannot fall below 0.66, as the golden answers in the KOLD dataset are determined by the majority vote of the three Korean annotators.

We also calculated the hate speech detection accuracy of each non-native participants who took part in the final pipeline validation experiment. The results are presented in Table \ref{tab:pipeline_participant_stats}. Every participant showed lower performance compared to the average accuracy of the Korean annotators. This implies the persistent gap between non-native moderators and native moderators. However, it is difficult to attribute the performance difference solely to the limitations of the non-native moderators.

The average accuracy of the Korean annotators was calculated across all samples in the KOLD dataset. In contrast, the accuracy of the pipeline validation experiment participants was measured on a filtered set of samples requiring cultural knowledge and understanding for proper moderation. This suggests that non-native moderators might perform better on the full dataset, as it includes samples that do not require cultural knowledge for moderation. Meanwhile, we did not assess the accuracy of native moderators using the same set of 28 samples as the non-native moderators. This isbecause 27 KOLD annotators participated in annotating those 28 samples, with all but one (who annotated two samples) working on only one sample. Calculating accuracy with 28 samples would result in each KOLD annotator having either 0\% or 100\% accuracy, making the averaged accuracy meaningless.

Thus, our results indicate that non-native moderators still fall short compared to native moderators. However, these findings should be interpreted with caution due to inherent limitations in the statistical comparison.

\subsection{Limitations of Early Decision-Making and Error Analysis}
\label{sec:quantitative_analysis_LLMerror}

\begin{table}[ht]
  \centering
  \begin{minipage}{0.8\columnwidth}
    \resizebox{\columnwidth}{!}{%
      \begin{tabular}{@{}lll@{}}
        \toprule
        & \multicolumn{2}{c}{LLM Moderators}                          \\ \cmidrule(l){2-3}
        KOLD Annotators & \multicolumn{1}{c}{Correct} & \multicolumn{1}{c}{Incorrect} \\ \midrule
        Agree           & 91 (0.63)                          & 14 (0.10)                            \\ %
        Disagree        & 21 (0.15)                          & 17 (0.12)                            \\ \bottomrule
      \end{tabular}%
    }
  \end{minipage}%
  \setlength{\belowdisplayskip}{-0.5em} %
  \[
    \chi^2 = 16.2064
  \]
  \setlength{\abovedisplayskip}{-0.5em}
  \[
    p = 0.000057 \, (< 0.05)
  \]
  \setlength{\belowdisplayskip}{-0.5em}
  \caption{Analysis of 143 samples that reached unanimous agreement in Step2 of our pipeline during the pipeline validation experiment. The samples were first categorized based on whether the LLM moderators' unanimous decision was correct. Then, the samples were divided according to the level of agreement among the three Korean annotators of the KOLD dataset. A Chi-square test was conducted,  decisions are significantly correlated with the agreement among the Korean annotators, reflecting the inherent difficulty of the samples.}
  \label{table:error_analysis}
\end{table}

While our pipeline effectively reduces human workload by leaveraging LLM moderators in step 2, it has certain limitations. In our pipeline, an early decision is made in Step 2 when all three LLM moderators reach a consensus. However, if they unanimously agree on an incorrect judgment at this stage, the pipeline lacks a mechanism to correct this error. In our pipeline validation experiment, 31 out of 143 early decision samples (18\% of all samples) resulted in incorrect unanimous agreements. In this discussion, we analyze the difficulty of these misclassified samples, as presented in Table \ref{table:error_analysis}.

We implicitly define the difficulty of a hate speech sample based on the agreement among native moderators. In the KOLD dataset, golden answers are determined by the majority vote of three annotators. If all three annotators agree, the sample is likely to be straightforward and reliable. Conversely, if the annotators disagree, the sample may be more ambiguous or challenging. To investigate the relationship between LLM agreement accuracy (143 samples) and the agreement level of KOLD human annotators, we conducted a Chi-square test to test the null hypothesis $H_0$: the accuracy of LLM-agree samples is independent of human agreement. The results showed a Chi-square value of 16.2064 and a p-value of 0.000057 (< 0.05), leading to the rejection of the null hypothesis. This indicates that the incorrect unanimous agreements in Step 2 are more likely to be inherently difficult even for native moderators. Thus, solving these samples may require a more advanced pipeline or the assistance of native moderators. The full sample analysis is in Appendix \ref{appendix:A_pipeline_sample_analysis}.

\section{Conclusion}
\label{sec:conclusion}
We presented \tool, a system that assists non-native moderators in cross-cultural hate speech detection through RAG-enhanced cultural context annotations and strategic human-LLM collaboration.
By addressing three key challenges identified from our user study---\textit{understanding culturally-specific knowledge}, \textit{navigating cultural sentiment differences}, and \textit{interpreting internet culture}---our system achieves 78\% accuracy while reducing human workload by 83.6\% in Korean hate speech moderation with Indonesian and German participants.
This demonstrates that non-native moderators, when supported with appropriate cultural context, can effectively contribute to content moderation across linguistic and cultural boundaries.
In future work, we aim to explore extending \tool to examine its effectiveness across different cultural and linguistic combinations, beyond the Korean-English pairing examined in our study.
We hope our findings contribute to advancing research in cross-cultural content moderation, addressing critical challenges in global online safety.

\section*{Limitations}
\label{sec:limitation}
\paragraph{Language Proxy Considerations}
The participants in our user study and pipeline evaluation experiment are from Indonesia and Germany, and English is not their first language. Thus, they relied on a proxy language (English) to understand the Korean content. This likely made it more challenging for them to fully grasp the nuances of the language when assessing the offensiveness of the content. To address this limitation, future work will involve translating the content into each participant's native language.

\paragraph{Early Decision-Making in the Pipeline} Our pipeline makes an early decision without additional offensiveness verification when the three LLM moderators reach an unanimous agreement. As a result, our pipeline cannot correct unanimous incorrect decisions made during the early decision stage. To minimize this risk, we selected 3 GPT-4o models since it is the combination which showed highest agree accuracy(Table \ref{tab:llms-accuracy}). Furthermore, errors that were not filtered out underwent quantitative analysis through Chi-square testing in Section \ref{sec:quantitative_analysis_LLMerror}, showing that the errors missed during early decisions in Step 2 in our pipeline were likely to involve more difficult cases or be inaccurate. However, since there remain cases where the LLMs make errors, future work should focus on addressing this limitation. Additionally, efforts to improve performance on challenging cases should also be prioritized. For example, increasing the number of LLM Moderators beyond the current three may enhance the reliability of the LLM uncertainty. Additionally, incorporating LLM consistency-checking methods alongside the use of LLM Moderators could further improve the robustness and accuracy of the system.

\section*{Ethical Considerations}
\label{sec:ethics}

\paragraph{Data}

Our study is conducted in a course project.
Each participant was paid 10,000 KRW, minimum wage.

\paragraph{Annotator Demographics}

All annotators were not native speakers of both the language (English) and culture (Korean) that they were annotating.
Other annotator demographics were not collected for this study, except for native language and nationality.

\paragraph{Compute/AI Resources}
All our experiments were conducted on local computers using API service.
The API calls to the GPT models were done through the Azure OpenAI service.
The Gemini model was accessed via the Google Gemini API service. The Claude model was accessed by Anthropic API service.
Finally, we also acknowledge the usage of ChatGPT and GitHub Copilot for building our codebase.

\section*{Acknowledgements}
\label{sec:ack}
This research project has benefitted from the Microsoft Accelerate Foundation Models Research (AFMR) grant program through which leading foundation models hosted by Microsoft Azure along with access to Azure credits were provided to conduct the research.
This work was supported by Institute of Information \& communications Technology Planning \& Evaluation (IITP) grant funded by the Korea government(MSIT) (No.RS-2022-II220184, Development and Study of AI Technologies to Inexpensively Conform to Evolving Policy on Ethics)


\begin{thebibliography}{47}
\providecommand{\natexlab}[1]{#1}

\bibitem[{Alkhatib and Bernstein(2019)}]{10.1145/3290605.3300760}
Ali Alkhatib and Michael Bernstein. 2019.
\newblock \href {https://doi.org/10.1145/3290605.3300760} {Street-level
  algorithms: A theory at the gaps between policy and decisions}.
\newblock In \emph{Proceedings of the 2019 CHI Conference on Human Factors in
  Computing Systems}, CHI '19, page 1–13, New York, NY, USA. Association for
  Computing Machinery.

\bibitem[{Arango~Monnar et~al.(2022)Arango~Monnar, Perez, Poblete, Salda{\~n}a,
  and Proust}]{arango-monnar-etal-2022-resources}
Ayme Arango~Monnar, Jorge Perez, Barbara Poblete, Magdalena Salda{\~n}a, and
  Valentina Proust. 2022.
\newblock \href {https://doi.org/10.18653/v1/2022.woah-1.12} {Resources for
  multilingual hate speech detection}.
\newblock In \emph{Proceedings of the Sixth Workshop on Online Abuse and Harms
  (WOAH)}, pages 122--130, Seattle, Washington (Hybrid). Association for
  Computational Linguistics.

\bibitem[{Breazu et~al.(2024)Breazu, Schirmer, Hu, and
  Katsos}]{breazu2024largelanguagemodelsthematic}
Petre Breazu, Miriam Schirmer, Songbo Hu, and Napoleon Katsos. 2024.
\newblock \href {https://arxiv.org/abs/2408.05126} {Large language models and
  thematic analysis: Human-ai synergy in researching hate speech on social
  media}.
\newblock \emph{Preprint}, arXiv:2408.05126.

\bibitem[{Chan et~al.(2024)Chan, Nguyen, and Joshi}]{chan2024hate}
Fai~Leui Chan, Duke Nguyen, and Aditya Joshi. 2024.
\newblock \href {https://arxiv.org/abs/2410.11230} {"is hate lost in
  translation?": Evaluation of multilingual lgbtqia+ hate speech detection}.
\newblock \emph{arXiv preprint arXiv:2410.11230}.

\bibitem[{Chung et~al.(2019)Chung, Kuzmenko, Tekiroglu, and
  Guerini}]{chung-etal-2019-conan}
Yi-Ling Chung, Elizaveta Kuzmenko, Serra~Sinem Tekiroglu, and Marco Guerini.
  2019.
\newblock \href {https://doi.org/10.18653/v1/P19-1271} {{CONAN} - {CO}unter
  {NA}rratives through nichesourcing: a multilingual dataset of responses to
  fight online hate speech}.
\newblock In \emph{Proceedings of the 57th Annual Meeting of the Association
  for Computational Linguistics}, pages 2819--2829, Florence, Italy.
  Association for Computational Linguistics.

\bibitem[{Demus et~al.(2022)Demus, Pitz, Sch{\"u}tz, Probol, Siegel, and
  Labudde}]{demus-etal-2022-comprehensive}
Christoph Demus, Jonas Pitz, Mina Sch{\"u}tz, Nadine Probol, Melanie Siegel,
  and Dirk Labudde. 2022.
\newblock \href {https://doi.org/10.18653/v1/2022.woah-1.14} {Detox: A
  comprehensive dataset for {G}erman offensive language and conversation
  analysis}.
\newblock In \emph{Proceedings of the Sixth Workshop on Online Abuse and Harms
  (WOAH)}, pages 143--153, Seattle, Washington (Hybrid). Association for
  Computational Linguistics.

\bibitem[{Deng et~al.(2022)Deng, Zhou, Sun, Zheng, Mi, Meng, and
  Huang}]{deng-etal-2022-cold}
Jiawen Deng, Jingyan Zhou, Hao Sun, Chujie Zheng, Fei Mi, Helen Meng, and
  Minlie Huang. 2022.
\newblock \href {https://doi.org/10.18653/v1/2022.emnlp-main.796} {{COLD}: A
  benchmark for {C}hinese offensive language detection}.
\newblock In \emph{Proceedings of the 2022 Conference on Empirical Methods in
  Natural Language Processing}, pages 11580--11599, Abu Dhabi, United Arab
  Emirates. Association for Computational Linguistics.

\bibitem[{Ding et~al.(2024)Ding, Ping, Gunturi, Carik, Stil, Wilhelm, Daryanto,
  Hawdon, Lee, and Rho}]{ding2024counterquillinvestigatingpotentialhumanai}
Xiaohan Ding, Kaike Ping, Uma~Sushmitha Gunturi, Buse Carik, Sophia Stil,
  Lance~T Wilhelm, Taufiq Daryanto, James Hawdon, Sang~Won Lee, and Eugenia~H
  Rho. 2024.
\newblock \href {https://arxiv.org/abs/2410.03032} {Counterquill: Investigating
  the potential of human-ai collaboration in online counterspeech writing}.
\newblock \emph{Preprint}, arXiv:2410.03032.

\bibitem[{ElSherief et~al.(2021)ElSherief, Ziems, Muchlinski, Anupindi,
  Seybolt, De~Choudhury, and Yang}]{elsherief-etal-2021-latent}
Mai ElSherief, Caleb Ziems, David Muchlinski, Vaishnavi Anupindi, Jordyn
  Seybolt, Munmun De~Choudhury, and Diyi Yang. 2021.
\newblock \href {https://doi.org/10.18653/v1/2021.emnlp-main.29} {Latent
  hatred: A benchmark for understanding implicit hate speech}.
\newblock In \emph{Proceedings of the 2021 Conference on Empirical Methods in
  Natural Language Processing}, pages 345--363, Online and Punta Cana,
  Dominican Republic. Association for Computational Linguistics.

\bibitem[{Elswah(2024)}]{elswah2024investigating}
Mona Elswah. 2024.
\newblock \href
  {https://cdt.org/research/investigating-content-moderation-systems-in-the-global-south}
  {Investigating content moderation systems in the global south}.
\newblock \emph{Center for Democracy and Technology}.

\bibitem[{Gorwa et~al.(2020)Gorwa, Binns, and
  Katzenbach}]{doi:10.1177/2053951719897945}
Robert Gorwa, Reuben Binns, and Christian Katzenbach. 2020.
\newblock \href {https://doi.org/10.1177/2053951719897945} {Algorithmic content
  moderation: Technical and political challenges in the automation of platform
  governance}.
\newblock \emph{Big Data \& Society}, 7(1):2053951719897945.

\bibitem[{Hee et~al.(2024)Hee, Sharma, Cao, Nandi, Nakov, Chakraborty, and
  Lee}]{hee-etal-2024-recent}
Ming~Shan Hee, Shivam Sharma, Rui Cao, Palash Nandi, Preslav Nakov, Tanmoy
  Chakraborty, and Roy Ka-Wei Lee. 2024.
\newblock \href {https://doi.org/10.18653/v1/2024.findings-emnlp.254} {Recent
  advances in online hate speech moderation: Multimodality and the role of
  large models}.
\newblock In \emph{Findings of the Association for Computational Linguistics:
  EMNLP 2024}, pages 4407--4419, Miami, Florida, USA. Association for
  Computational Linguistics.

\bibitem[{Jeong et~al.(2022)Jeong, Oh, Lee, Ahn, Moon, Park, and
  Oh}]{jeong-etal-2022-kold}
Younghoon Jeong, Juhyun Oh, Jongwon Lee, Jaimeen Ahn, Jihyung Moon, Sungjoon
  Park, and Alice Oh. 2022.
\newblock \href {https://doi.org/10.18653/v1/2022.emnlp-main.744} {{KOLD}:
  {K}orean offensive language dataset}.
\newblock In \emph{Proceedings of the 2022 Conference on Empirical Methods in
  Natural Language Processing}, pages 10818--10833, Abu Dhabi, United Arab
  Emirates. Association for Computational Linguistics.

\bibitem[{Jha et~al.(2024)Jha, Jain, Mandal, Chadha, Saha, and
  Bhattacharyya}]{jha-etal-2024-memeguard}
Prince Jha, Raghav Jain, Konika Mandal, Aman Chadha, Sriparna Saha, and Pushpak
  Bhattacharyya. 2024.
\newblock \href {https://doi.org/10.18653/v1/2024.acl-long.439} {{M}eme{G}uard:
  An {LLM} and {VLM}-based framework for advancing content moderation via meme
  intervention}.
\newblock In \emph{Proceedings of the 62nd Annual Meeting of the Association
  for Computational Linguistics (Volume 1: Long Papers)}, pages 8084--8104,
  Bangkok, Thailand. Association for Computational Linguistics.

\bibitem[{Jhaver et~al.(2019)Jhaver, Birman, Gilbert, and
  Bruckman}]{10.1145/3338243}
Shagun Jhaver, Iris Birman, Eric Gilbert, and Amy Bruckman. 2019.
\newblock \href {https://doi.org/10.1145/3338243} {Human-machine collaboration
  for content regulation: The case of reddit automoderator}.
\newblock \emph{ACM Trans. Comput.-Hum. Interact.}, 26(5).

\bibitem[{Jin et~al.(2024)Jin, Kim, Lee, Yoo, Oh, and
  Lee}]{jin-etal-2024-kobbq}
Jiho Jin, Jiseon Kim, Nayeon Lee, Haneul Yoo, Alice Oh, and Hwaran Lee. 2024.
\newblock \href {https://doi.org/10.1162/tacl_a_00661} {{K}o{BBQ}: {K}orean
  bias benchmark for question answering}.
\newblock \emph{Transactions of the Association for Computational Linguistics},
  12:507--524.

\bibitem[{Kang and Qian(2024)}]{kang-qian-2024-implanting}
Hankun Kang and Tieyun Qian. 2024.
\newblock \href {https://doi.org/10.18653/v1/2024.findings-acl.56} {Implanting
  {LLM}`s knowledge via reading comprehension tree for toxicity detection}.
\newblock In \emph{Findings of the Association for Computational Linguistics:
  ACL 2024}, pages 947--962, Bangkok, Thailand. Association for Computational
  Linguistics.

\bibitem[{Kolla et~al.(2024{\natexlab{a}})Kolla, Salunkhe, Chandrasekharan, and
  Saha}]{kolla2024llm}
Mahi Kolla, Siddharth Salunkhe, Eshwar Chandrasekharan, and Koustuv Saha.
  2024{\natexlab{a}}.
\newblock \href {https://doi.org/10.1145/3613905.3650828} {Llm-mod: Can large
  language models assist content moderation?}
\newblock In \emph{Extended Abstracts of the CHI Conference on Human Factors in
  Computing Systems}, CHI EA '24, New York, NY, USA. Association for Computing
  Machinery.

\bibitem[{Kolla et~al.(2024{\natexlab{b}})Kolla, Salunkhe, Chandrasekharan, and
  Saha}]{10.1145/3613905.3650828}
Mahi Kolla, Siddharth Salunkhe, Eshwar Chandrasekharan, and Koustuv Saha.
  2024{\natexlab{b}}.
\newblock \href {https://doi.org/10.1145/3613905.3650828} {Llm-mod: Can large
  language models assist content moderation?}
\newblock In \emph{Extended Abstracts of the CHI Conference on Human Factors in
  Computing Systems}, CHI EA '24, New York, NY, USA. Association for Computing
  Machinery.

\bibitem[{Kumar et~al.(2024)Kumar, AbuHashem, and
  Durumeric}]{Kumar_AbuHashem_Durumeric_2024}
Deepak Kumar, Yousef~Anees AbuHashem, and Zakir Durumeric. 2024.
\newblock \href {https://doi.org/10.1609/icwsm.v18i1.31358} {Watch your
  language: Investigating content moderation with large language models}.
\newblock \emph{Proceedings of the International AAAI Conference on Web and
  Social Media}, 18(1):865--878.

\bibitem[{Lee et~al.(2024)Lee, Jung, Myung, Jin, Camacho-Collados, Kim, and
  Oh}]{lee-etal-2024-exploring-cross}
Nayeon Lee, Chani Jung, Junho Myung, Jiho Jin, Jose Camacho-Collados, Juho Kim,
  and Alice Oh. 2024.
\newblock \href {https://doi.org/10.18653/v1/2024.naacl-long.236} {Exploring
  cross-cultural differences in {E}nglish hate speech annotations: From dataset
  construction to analysis}.
\newblock In \emph{Proceedings of the 2024 Conference of the North American
  Chapter of the Association for Computational Linguistics: Human Language
  Technologies (Volume 1: Long Papers)}, pages 4205--4224, Mexico City, Mexico.
  Association for Computational Linguistics.

\bibitem[{Lee et~al.(2023)Lee, Jung, and Oh}]{lee-etal-2023-hate}
Nayeon Lee, Chani Jung, and Alice Oh. 2023.
\newblock \href {https://doi.org/10.18653/v1/2023.c3nlp-1.5} {Hate speech
  classifiers are culturally insensitive}.
\newblock In \emph{Proceedings of the First Workshop on Cross-Cultural
  Considerations in NLP (C3NLP)}, pages 35--46, Dubrovnik, Croatia. Association
  for Computational Linguistics.

\bibitem[{Lewis et~al.(2020)Lewis, Perez, Piktus, Petroni, Karpukhin, Goyal,
  K\"{u}ttler, Lewis, Yih, Rockt\"{a}schel, Riedel, and
  Kiela}]{lewis2020retrieval}
Patrick Lewis, Ethan Perez, Aleksandra Piktus, Fabio Petroni, Vladimir
  Karpukhin, Naman Goyal, Heinrich K\"{u}ttler, Mike Lewis, Wen-tau Yih, Tim
  Rockt\"{a}schel, Sebastian Riedel, and Douwe Kiela. 2020.
\newblock \href {https://doi.org/10.5555/3495724.3496517} {Retrieval-augmented
  generation for knowledge-intensive nlp tasks}.
\newblock In \emph{Proceedings of the 34th International Conference on Neural
  Information Processing Systems}, NIPS '20, Red Hook, NY, USA. Curran
  Associates Inc.

\bibitem[{Li et~al.(2024)Li, Teney, Yang, Wen, Xie, and
  Wang}]{li2024cultureparkboostingcrossculturalunderstanding}
Cheng Li, Damien Teney, Linyi Yang, Qingsong Wen, Xing Xie, and Jindong Wang.
  2024.
\newblock \href {https://arxiv.org/abs/2405.15145} {Culturepark: Boosting
  cross-cultural understanding in large language models}.
\newblock \emph{Preprint}, arXiv:2405.15145.

\bibitem[{Maronikolakis et~al.(2022)Maronikolakis, Wisiorek, Nann, Jabbar,
  Udupa, and Schuetze}]{maronikolakis-etal-2022-listening}
Antonis Maronikolakis, Axel Wisiorek, Leah Nann, Haris Jabbar, Sahana Udupa,
  and Hinrich Schuetze. 2022.
\newblock \href {https://doi.org/10.18653/v1/2022.findings-acl.87} {Listening
  to affected communities to define extreme speech: Dataset and experiments}.
\newblock In \emph{Findings of the Association for Computational Linguistics:
  ACL 2022}, pages 1089--1104, Dublin, Ireland. Association for Computational
  Linguistics.

\bibitem[{Masud et~al.(2022)Masud, Bedi, Khan, Akhtar, and
  Chakraborty}]{10.1145/3534678.3539161}
Sarah Masud, Manjot Bedi, Mohammad~Aflah Khan, Md~Shad Akhtar, and Tanmoy
  Chakraborty. 2022.
\newblock \href {https://doi.org/10.1145/3534678.3539161} {Proactively reducing
  the hate intensity of online posts via hate speech normalization}.
\newblock In \emph{Proceedings of the 28th ACM SIGKDD Conference on Knowledge
  Discovery and Data Mining}, KDD '22, page 3524–3534, New York, NY, USA.
  Association for Computing Machinery.

\bibitem[{Masud et~al.(2024)Masud, Singh, Hangya, Fraser, and
  Chakraborty}]{masud-etal-2024-hate}
Sarah Masud, Sahajpreet Singh, Viktor Hangya, Alexander Fraser, and Tanmoy
  Chakraborty. 2024.
\newblock \href {https://doi.org/10.18653/v1/2024.emnlp-main.886} {Hate
  personified: Investigating the role of {LLM}s in content moderation}.
\newblock In \emph{Proceedings of the 2024 Conference on Empirical Methods in
  Natural Language Processing}, pages 15847--15863, Miami, Florida, USA.
  Association for Computational Linguistics.

\bibitem[{Mathew et~al.(2021)Mathew, Saha, Yimam, Biemann, Goyal, and
  Mukherjee}]{Mathew_Saha_Yimam_Biemann_Goyal_Mukherjee_2021}
Binny Mathew, Punyajoy Saha, Seid~Muhie Yimam, Chris Biemann, Pawan Goyal, and
  Animesh Mukherjee. 2021.
\newblock \href {https://doi.org/10.1609/aaai.v35i17.17745} {Hatexplain: A
  benchmark dataset for explainable hate speech detection}.
\newblock \emph{Proceedings of the AAAI Conference on Artificial Intelligence},
  35(17):14867--14875.

\bibitem[{Milmo(2021)}]{milmo2021facebook}
Dan Milmo. 2021.
\newblock \href
  {https://www.theguardian.com/technology/2021/oct/25/facebook-revelations-from-misinformation-to-mental-health}
  {Facebook revelations: what is in cache of internal documents?}
\newblock \emph{The Guardian}.

\bibitem[{Muhammad et~al.(2025)Muhammad, Abdulmumin, Ayele, Adelani, Ahmad,
  Aliyu, Onyango, Wanzare, Rutunda, Aliyu, Alemneh, Hourrane, Gebremichael,
  Ismail, Beloucif, Jibril, Bukula, Mabuya, Osei, Oppong, Belay, Guge, Asfaw,
  Chukwuneke, Röttger, Yimam, and
  Ousidhoum}]{muhammad2025afrihatemultilingualcollectionhate}
Shamsuddeen~Hassan Muhammad, Idris Abdulmumin, Abinew~Ali Ayele, David~Ifeoluwa
  Adelani, Ibrahim~Said Ahmad, Saminu~Mohammad Aliyu, Nelson~Odhiambo Onyango,
  Lilian D.~A. Wanzare, Samuel Rutunda, Lukman~Jibril Aliyu, Esubalew Alemneh,
  Oumaima Hourrane, Hagos~Tesfahun Gebremichael, Elyas~Abdi Ismail, Meriem
  Beloucif, Ebrahim~Chekol Jibril, Andiswa Bukula, Rooweither Mabuya, Salomey
  Osei, Abigail Oppong, Tadesse~Destaw Belay, Tadesse~Kebede Guge,
  Tesfa~Tegegne Asfaw, Chiamaka~Ijeoma Chukwuneke, Paul Röttger, Seid~Muhie
  Yimam, and Nedjma Ousidhoum. 2025.
\newblock \href {https://arxiv.org/abs/2501.08284} {Afrihate: A multilingual
  collection of hate speech and abusive language datasets for african
  languages}.
\newblock \emph{Preprint}, arXiv:2501.08284.

\bibitem[{Nigatu and Raji(2024)}]{nigatu2024searched}
Hellina~Hailu Nigatu and Inioluwa~Deborah Raji. 2024.
\newblock \href {https://doi.org/10.1145/3630106.3658546} {“i searched for a
  religious song in amharic and got sexual content instead’’: Investigating
  online harm in low-resourced languages on youtube.}
\newblock In \emph{Proceedings of the 2024 ACM Conference on Fairness,
  Accountability, and Transparency}, FAccT '24, page 141–160, New York, NY,
  USA. Association for Computing Machinery.

\bibitem[{OpenAI et~al.(2024)OpenAI, :, Hurst, Lerer, Goucher, Perelman,
  Ramesh, Clark, Ostrow, Welihinda, Hayes, Radford, Mądry, Baker-Whitcomb,
  Beutel, Borzunov, Carney, Chow, Kirillov, Nichol, Paino, Renzin, Passos,
  Kirillov, Christakis, Conneau, Kamali, Jabri, Moyer, Tam, Crookes,
  Tootoochian, Tootoonchian, Kumar, Vallone, Karpathy, Braunstein, Cann,
  Codispoti, Galu, Kondrich, Tulloch, Mishchenko, Baek, Jiang, Pelisse,
  Woodford, Gosalia, Dhar, Pantuliano, Nayak, Oliver, Zoph, Ghorbani,
  Leimberger, Rossen, Sokolowsky, Wang, Zweig, Hoover, Samic, McGrew, Spero,
  Giertler, Cheng, Lightcap, Walkin, Quinn, Guarraci, Hsu, Kellogg, Eastman,
  Lugaresi, Wainwright, Bassin, Hudson, Chu, Nelson, Li, Shern, Conger,
  Barette, Voss, Ding, Lu, Zhang, Beaumont, Hallacy, Koch, Gibson, Kim, Choi,
  McLeavey, Hesse, Fischer, Winter, Czarnecki, Jarvis, Wei, Koumouzelis,
  Sherburn, Kappler, Levin, Levy, Carr, Farhi, Mely, Robinson, Sasaki, Jin,
  Valladares, Tsipras, Li, Nguyen, Findlay, Oiwoh, Wong, Asdar, Proehl, Yang,
  Antonow, Kramer, Peterson, Sigler, Wallace, Brevdo, Mays, Khorasani, Such,
  Raso, Zhang, von Lohmann, Sulit, Goh, Oden, Salmon, Starace, Brockman,
  Salman, Bao, Hu, Wong, Wang, Schmidt, Whitney, Jun, Kirchner,
  de~Oliveira~Pinto, Ren, Chang, Chung, Kivlichan, O'Connell, O'Connell,
  Osband, Silber, Sohl, Okuyucu, Lan, Kostrikov, Sutskever, Kanitscheider,
  Gulrajani, Coxon, Menick, Pachocki, Aung, Betker, Crooks, Lennon, Kiros,
  Leike, Park, Kwon, Phang, Teplitz, Wei, Wolfe, Chen, Harris, Varavva, Lee,
  Shieh, Lin, Yu, Weng, Tang, Yu, Jang, Candela, Beutler, Landers, Parish,
  Heidecke, Schulman, Lachman, McKay, Uesato, Ward, Kim, Huizinga, Sitkin,
  Kraaijeveld, Gross, Kaplan, Snyder, Achiam, Jiao, Lee, Zhuang, Harriman,
  Fricke, Hayashi, Singhal, Shi, Karthik, Wood, Rimbach, Hsu, Nguyen,
  Gu-Lemberg, Button, Liu, Howe, Muthukumar, Luther, Ahmad, Kai, Itow, Workman,
  Pathak, Chen, Jing, Guy, Fedus, Zhou, Mamitsuka, Weng, McCallum, Held,
  Ouyang, Feuvrier, Zhang, Kondraciuk, Kaiser, Hewitt, Metz, Doshi, Aflak,
  Simens, Boyd, Thompson, Dukhan, Chen, Gray, Hudnall, Zhang, Aljubeh, Litwin,
  Zeng, Johnson, Shetty, Gupta, Shah, Yatbaz, Yang, Zhong, Glaese, Chen,
  Janner, Lampe, Petrov, Wu, Wang, Fradin, Pokrass, Castro, de~Castro, Pavlov,
  Brundage, Wang, Khan, Murati, Bavarian, Lin, Yesildal, Soto, Gimelshein,
  Cone, Staudacher, Summers, LaFontaine, Chowdhury, Ryder, Stathas, Turley,
  Tezak, Felix, Kudige, Keskar, Deutsch, Bundick, Puckett, Nachum, Okelola,
  Boiko, Murk, Jaffe, Watkins, Godement, Campbell-Moore, Chao, McMillan, Belov,
  Su, Bak, Bakkum, Deng, Dolan, Hoeschele, Welinder, Tillet, Pronin, Tillet,
  Dhariwal, Yuan, Dias, Lim, Arora, Troll, Lin, Lopes, Puri, Miyara, Leike,
  Gaubert, Zamani, Wang, Donnelly, Honsby, Smith, Sahai, Ramchandani, Huet,
  Carmichael, Zellers, Chen, Chen, Nigmatullin, Cheu, Jain, Altman, Schoenholz,
  Toizer, Miserendino, Agarwal, Culver, Ethersmith, Gray, Grove, Metzger,
  Hermani, Jain, Zhao, Wu, Jomoto, Wu, Shuaiqi, Xia, Phene, Papay, Narayanan,
  Coffey, Lee, Hall, Balaji, Broda, Stramer, Xu, Gogineni, Christianson,
  Sanders, Patwardhan, Cunninghman, Degry, Dimson, Raoux, Shadwell, Zheng,
  Underwood, Markov, Sherbakov, Rubin, Stasi, Kaftan, Heywood, Peterson,
  Walters, Eloundou, Qi, Moeller, Monaco, Kuo, Fomenko, Chang, Zheng, Zhou,
  Manassra, Sheu, Zaremba, Patil, Qian, Kim, Cheng, Zhang, He, Zhang, Jin, Dai,
  and Malkov}]{hurst2024gpt}
OpenAI, :, Aaron Hurst, Adam Lerer, Adam~P. Goucher, Adam Perelman, Aditya
  Ramesh, Aidan Clark, AJ~Ostrow, Akila Welihinda, Alan Hayes, Alec Radford,
  Aleksander Mądry, Alex Baker-Whitcomb, Alex Beutel, Alex Borzunov, Alex
  Carney, Alex Chow, Alex Kirillov, Alex Nichol, Alex Paino, Alex Renzin,
  Alex~Tachard Passos, Alexander Kirillov, Alexi Christakis, Alexis Conneau,
  Ali Kamali, Allan Jabri, Allison Moyer, Allison Tam, Amadou Crookes, Amin
  Tootoochian, Amin Tootoonchian, Ananya Kumar, Andrea Vallone, Andrej
  Karpathy, Andrew Braunstein, Andrew Cann, Andrew Codispoti, Andrew Galu,
  Andrew Kondrich, Andrew Tulloch, Andrey Mishchenko, Angela Baek, Angela
  Jiang, Antoine Pelisse, Antonia Woodford, Anuj Gosalia, Arka Dhar, Ashley
  Pantuliano, Avi Nayak, Avital Oliver, Barret Zoph, Behrooz Ghorbani, Ben
  Leimberger, Ben Rossen, Ben Sokolowsky, Ben Wang, Benjamin Zweig, Beth
  Hoover, Blake Samic, Bob McGrew, Bobby Spero, Bogo Giertler, Bowen Cheng,
  Brad Lightcap, Brandon Walkin, Brendan Quinn, Brian Guarraci, Brian Hsu,
  Bright Kellogg, Brydon Eastman, Camillo Lugaresi, Carroll Wainwright, Cary
  Bassin, Cary Hudson, Casey Chu, Chad Nelson, Chak Li, Chan~Jun Shern,
  Channing Conger, Charlotte Barette, Chelsea Voss, Chen Ding, Cheng Lu, Chong
  Zhang, Chris Beaumont, Chris Hallacy, Chris Koch, Christian Gibson, Christina
  Kim, Christine Choi, Christine McLeavey, Christopher Hesse, Claudia Fischer,
  Clemens Winter, Coley Czarnecki, Colin Jarvis, Colin Wei, Constantin
  Koumouzelis, Dane Sherburn, Daniel Kappler, Daniel Levin, Daniel Levy, David
  Carr, David Farhi, David Mely, David Robinson, David Sasaki, Denny Jin, Dev
  Valladares, Dimitris Tsipras, Doug Li, Duc~Phong Nguyen, Duncan Findlay,
  Edede Oiwoh, Edmund Wong, Ehsan Asdar, Elizabeth Proehl, Elizabeth Yang, Eric
  Antonow, Eric Kramer, Eric Peterson, Eric Sigler, Eric Wallace, Eugene
  Brevdo, Evan Mays, Farzad Khorasani, Felipe~Petroski Such, Filippo Raso,
  Francis Zhang, Fred von Lohmann, Freddie Sulit, Gabriel Goh, Gene Oden, Geoff
  Salmon, Giulio Starace, Greg Brockman, Hadi Salman, Haiming Bao, Haitang Hu,
  Hannah Wong, Haoyu Wang, Heather Schmidt, Heather Whitney, Heewoo Jun,
  Hendrik Kirchner, Henrique~Ponde de~Oliveira~Pinto, Hongyu Ren, Huiwen Chang,
  Hyung~Won Chung, Ian Kivlichan, Ian O'Connell, Ian O'Connell, Ian Osband, Ian
  Silber, Ian Sohl, Ibrahim Okuyucu, Ikai Lan, Ilya Kostrikov, Ilya Sutskever,
  Ingmar Kanitscheider, Ishaan Gulrajani, Jacob Coxon, Jacob Menick, Jakub
  Pachocki, James Aung, James Betker, James Crooks, James Lennon, Jamie Kiros,
  Jan Leike, Jane Park, Jason Kwon, Jason Phang, Jason Teplitz, Jason Wei,
  Jason Wolfe, Jay Chen, Jeff Harris, Jenia Varavva, Jessica~Gan Lee, Jessica
  Shieh, Ji~Lin, Jiahui Yu, Jiayi Weng, Jie Tang, Jieqi Yu, Joanne Jang,
  Joaquin~Quinonero Candela, Joe Beutler, Joe Landers, Joel Parish, Johannes
  Heidecke, John Schulman, Jonathan Lachman, Jonathan McKay, Jonathan Uesato,
  Jonathan Ward, Jong~Wook Kim, Joost Huizinga, Jordan Sitkin, Jos Kraaijeveld,
  Josh Gross, Josh Kaplan, Josh Snyder, Joshua Achiam, Joy Jiao, Joyce Lee,
  Juntang Zhuang, Justyn Harriman, Kai Fricke, Kai Hayashi, Karan Singhal, Katy
  Shi, Kavin Karthik, Kayla Wood, Kendra Rimbach, Kenny Hsu, Kenny Nguyen,
  Keren Gu-Lemberg, Kevin Button, Kevin Liu, Kiel Howe, Krithika Muthukumar,
  Kyle Luther, Lama Ahmad, Larry Kai, Lauren Itow, Lauren Workman, Leher
  Pathak, Leo Chen, Li~Jing, Lia Guy, Liam Fedus, Liang Zhou, Lien Mamitsuka,
  Lilian Weng, Lindsay McCallum, Lindsey Held, Long Ouyang, Louis Feuvrier,
  Lu~Zhang, Lukas Kondraciuk, Lukasz Kaiser, Luke Hewitt, Luke Metz, Lyric
  Doshi, Mada Aflak, Maddie Simens, Madelaine Boyd, Madeleine Thompson, Marat
  Dukhan, Mark Chen, Mark Gray, Mark Hudnall, Marvin Zhang, Marwan Aljubeh,
  Mateusz Litwin, Matthew Zeng, Max Johnson, Maya Shetty, Mayank Gupta, Meghan
  Shah, Mehmet Yatbaz, Meng~Jia Yang, Mengchao Zhong, Mia Glaese, Mianna Chen,
  Michael Janner, Michael Lampe, Michael Petrov, Michael Wu, Michele Wang,
  Michelle Fradin, Michelle Pokrass, Miguel Castro, Miguel Oom~Temudo
  de~Castro, Mikhail Pavlov, Miles Brundage, Miles Wang, Minal Khan, Mira
  Murati, Mo~Bavarian, Molly Lin, Murat Yesildal, Nacho Soto, Natalia
  Gimelshein, Natalie Cone, Natalie Staudacher, Natalie Summers, Natan
  LaFontaine, Neil Chowdhury, Nick Ryder, Nick Stathas, Nick Turley, Nik Tezak,
  Niko Felix, Nithanth Kudige, Nitish Keskar, Noah Deutsch, Noel Bundick, Nora
  Puckett, Ofir Nachum, Ola Okelola, Oleg Boiko, Oleg Murk, Oliver Jaffe,
  Olivia Watkins, Olivier Godement, Owen Campbell-Moore, Patrick Chao, Paul
  McMillan, Pavel Belov, Peng Su, Peter Bak, Peter Bakkum, Peter Deng, Peter
  Dolan, Peter Hoeschele, Peter Welinder, Phil Tillet, Philip Pronin, Philippe
  Tillet, Prafulla Dhariwal, Qiming Yuan, Rachel Dias, Rachel Lim, Rahul Arora,
  Rajan Troll, Randall Lin, Rapha~Gontijo Lopes, Raul Puri, Reah Miyara, Reimar
  Leike, Renaud Gaubert, Reza Zamani, Ricky Wang, Rob Donnelly, Rob Honsby,
  Rocky Smith, Rohan Sahai, Rohit Ramchandani, Romain Huet, Rory Carmichael,
  Rowan Zellers, Roy Chen, Ruby Chen, Ruslan Nigmatullin, Ryan Cheu, Saachi
  Jain, Sam Altman, Sam Schoenholz, Sam Toizer, Samuel Miserendino, Sandhini
  Agarwal, Sara Culver, Scott Ethersmith, Scott Gray, Sean Grove, Sean Metzger,
  Shamez Hermani, Shantanu Jain, Shengjia Zhao, Sherwin Wu, Shino Jomoto,
  Shirong Wu, Shuaiqi, Xia, Sonia Phene, Spencer Papay, Srinivas Narayanan,
  Steve Coffey, Steve Lee, Stewart Hall, Suchir Balaji, Tal Broda, Tal Stramer,
  Tao Xu, Tarun Gogineni, Taya Christianson, Ted Sanders, Tejal Patwardhan,
  Thomas Cunninghman, Thomas Degry, Thomas Dimson, Thomas Raoux, Thomas
  Shadwell, Tianhao Zheng, Todd Underwood, Todor Markov, Toki Sherbakov, Tom
  Rubin, Tom Stasi, Tomer Kaftan, Tristan Heywood, Troy Peterson, Tyce Walters,
  Tyna Eloundou, Valerie Qi, Veit Moeller, Vinnie Monaco, Vishal Kuo, Vlad
  Fomenko, Wayne Chang, Weiyi Zheng, Wenda Zhou, Wesam Manassra, Will Sheu,
  Wojciech Zaremba, Yash Patil, Yilei Qian, Yongjik Kim, Youlong Cheng,
  Yu~Zhang, Yuchen He, Yuchen Zhang, Yujia Jin, Yunxing Dai, and Yury Malkov.
  2024.
\newblock \href {https://arxiv.org/abs/2410.21276} {Gpt-4o system card}.

\bibitem[{Park and Fung(2017)}]{park-fung-2017-one}
Ji~Ho Park and Pascale Fung. 2017.
\newblock \href {https://doi.org/10.18653/v1/W17-3006} {One-step and two-step
  classification for abusive language detection on {T}witter}.
\newblock In \emph{Proceedings of the First Workshop on Abusive Language
  Online}, pages 41--45, Vancouver, BC, Canada. Association for Computational
  Linguistics.

\bibitem[{Pawar et~al.(2024)Pawar, Park, Jin, Arora, Myung, Yadav, Haznitrama,
  Song, Oh, and Augenstein}]{pawar2024survey}
Siddhesh Pawar, Junyeong Park, Jiho Jin, Arnav Arora, Junho Myung, Srishti
  Yadav, Faiz~Ghifari Haznitrama, Inhwa Song, Alice Oh, and Isabelle
  Augenstein. 2024.
\newblock \href {https://arxiv.org/abs/2411.00860} {Survey of cultural
  awareness in language models: Text and beyond}.
\newblock \emph{arXiv preprint arXiv:2411.00860}.

\bibitem[{Podolak et~al.(2024)Podolak, {\L}ukasik, Balawender, Ossowski,
  Piotrowski, Bakowicz, and Sankowski}]{podolak-etal-2024-llm}
Jakub Podolak, Szymon {\L}ukasik, Pawe{\l} Balawender, Jan Ossowski, Jan
  Piotrowski, Katarzyna Bakowicz, and Piotr Sankowski. 2024.
\newblock \href {https://doi.org/10.18653/v1/2024.findings-emnlp.931} {{LLM}
  generated responses to mitigate the impact of hate speech}.
\newblock In \emph{Findings of the Association for Computational Linguistics:
  EMNLP 2024}, pages 15860--15876, Miami, Florida, USA. Association for
  Computational Linguistics.

\bibitem[{Roy et~al.(2023)Roy, Harshvardhan, Mukherjee, and
  Saha}]{roy-etal-2023-probing}
Sarthak Roy, Ashish Harshvardhan, Animesh Mukherjee, and Punyajoy Saha. 2023.
\newblock \href {https://doi.org/10.18653/v1/2023.findings-emnlp.407} {Probing
  {LLM}s for hate speech detection: strengths and vulnerabilities}.
\newblock In \emph{Findings of the Association for Computational Linguistics:
  EMNLP 2023}, pages 6116--6128, Singapore. Association for Computational
  Linguistics.

\bibitem[{Sap et~al.(2020)Sap, Gabriel, Qin, Jurafsky, Smith, and
  Choi}]{sap-etal-2020-social}
Maarten Sap, Saadia Gabriel, Lianhui Qin, Dan Jurafsky, Noah~A. Smith, and
  Yejin Choi. 2020.
\newblock \href {https://doi.org/10.18653/v1/2020.acl-main.486} {Social bias
  frames: Reasoning about social and power implications of language}.
\newblock In \emph{Proceedings of the 58th Annual Meeting of the Association
  for Computational Linguistics}, pages 5477--5490, Online. Association for
  Computational Linguistics.

\bibitem[{Thomas et~al.(2024)Thomas, Kelley, Tao, Meiklejohn, Vallis, Tan,
  Bratanič, Ferreira, Eranti, and
  Bursztein}]{thomas2024supportinghumanratersdetection}
Kurt Thomas, Patrick~Gage Kelley, David Tao, Sarah Meiklejohn, Owen Vallis,
  Shunwen Tan, Blaž Bratanič, Felipe~Tiengo Ferreira, Vijay~Kumar Eranti, and
  Elie Bursztein. 2024.
\newblock \href {https://arxiv.org/abs/2406.12800} {Supporting human raters
  with the detection of harmful content using large language models}.
\newblock \emph{Preprint}, arXiv:2406.12800.

\bibitem[{Vidgen et~al.(2021)Vidgen, Thrush, Waseem, and
  Kiela}]{vidgen-etal-2021-learning}
Bertie Vidgen, Tristan Thrush, Zeerak Waseem, and Douwe Kiela. 2021.
\newblock \href {https://doi.org/10.18653/v1/2021.acl-long.132} {Learning from
  the worst: Dynamically generated datasets to improve online hate detection}.
\newblock In \emph{Proceedings of the 59th Annual Meeting of the Association
  for Computational Linguistics and the 11th International Joint Conference on
  Natural Language Processing (Volume 1: Long Papers)}, pages 1667--1682,
  Online. Association for Computational Linguistics.

\bibitem[{Vishwamitra et~al.(2024)Vishwamitra, Guo, Romit, Ondracek, Cheng,
  Zhao, and Hu}]{10646809}
Nishant Vishwamitra, Keyan Guo, Farhan~Tajwar Romit, Isabelle Ondracek, Long
  Cheng, Ziming Zhao, and Hongxin Hu. 2024.
\newblock \href {https://doi.org/10.1109/SP54263.2024.00181} {Moderating new
  waves of online hate with chain-of-thought reasoning in large language
  models}.
\newblock In \emph{2024 IEEE Symposium on Security and Privacy (SP)}, pages
  788--806.

\bibitem[{Wang et~al.(2023)Wang, Hee, Awal, Choo, and
  Lee}]{10.24963/ijcai.2023/694}
Han Wang, Ming~Shan Hee, Md~Rabiul Awal, Kenny Tsu~Wei Choo, and Roy Ka-Wei
  Lee. 2023.
\newblock \href {https://doi.org/10.24963/ijcai.2023/694} {Evaluating gpt-3
  generated explanations for hateful content moderation}.
\newblock In \emph{Proceedings of the Thirty-Second International Joint
  Conference on Artificial Intelligence}, IJCAI '23.

\bibitem[{Wei et~al.(2022)Wei, Wang, Schuurmans, Bosma, ichter, Xia, Chi, Le,
  and Zhou}]{NEURIPS2022_9d560961}
Jason Wei, Xuezhi Wang, Dale Schuurmans, Maarten Bosma, brian ichter, Fei Xia,
  Ed~Chi, Quoc~V Le, and Denny Zhou. 2022.
\newblock \href
  {https://proceedings.neurips.cc/paper_files/paper/2022/file/9d5609613524ecf4f15af0f7b31abca4-Paper-Conference.pdf}
  {Chain-of-thought prompting elicits reasoning in large language models}.
\newblock In \emph{Advances in Neural Information Processing Systems},
  volume~35, pages 24824--24837. Curran Associates, Inc.

\bibitem[{Witness(2023)}]{witness2023big}
Global Witness. 2023.
\newblock \href
  {https://www.globalwitness.org/en/campaigns/digital-threats/how-big-tech-platforms-are-neglecting-their-non-english-language-users}
  {How {Big Tech} platforms are neglecting their non-{English} language users}.
\newblock \emph{Global Witness Org}.
\newblock Investigative report analyzing social media platforms' content
  moderation resources across different languages in the EU, based on Digital
  Service Act transparency reports.

\bibitem[{Xu et~al.(2024)Xu, Jain, and Kankanhalli}]{xu2024hallucination}
Ziwei Xu, Sanjay Jain, and Mohan Kankanhalli. 2024.
\newblock Hallucination is inevitable: An innate limitation of large language
  models.
\newblock \emph{arXiv preprint arXiv:2401.11817}.

\bibitem[{Yang et~al.(2023)Yang, Kim, Kim, Ho, Thorne, and
  Yun}]{yang-etal-2023-hare}
Yongjin Yang, Joonkee Kim, Yujin Kim, Namgyu Ho, James Thorne, and Se-Young
  Yun. 2023.
\newblock \href {https://doi.org/10.18653/v1/2023.findings-emnlp.365} {{HARE}:
  Explainable hate speech detection with step-by-step reasoning}.
\newblock In \emph{Findings of the Association for Computational Linguistics:
  EMNLP 2023}, pages 5490--5505, Singapore. Association for Computational
  Linguistics.

\bibitem[{Ye et~al.(2024)Ye, Wisiorek, Maronikolakis, Özge Alaçam, and
  Schütze}]{ye2024federatedapproachfewshothate}
Haotian Ye, Axel Wisiorek, Antonis Maronikolakis, Özge Alaçam, and Hinrich
  Schütze. 2024.
\newblock \href {https://arxiv.org/abs/2412.04942} {A federated approach to
  few-shot hate speech detection for marginalized communities}.
\newblock \emph{Preprint}, arXiv:2412.04942.

\bibitem[{Zhang et~al.(2024)Zhang, He, Ji, and Lu}]{zhang-etal-2024-dont-go}
Min Zhang, Jianfeng He, Taoran Ji, and Chang-Tien Lu. 2024.
\newblock \href {https://doi.org/10.18653/v1/2024.acl-long.652} {Don`t go to
  extremes: Revealing the excessive sensitivity and calibration limitations of
  {LLM}s in implicit hate speech detection}.
\newblock In \emph{Proceedings of the 62nd Annual Meeting of the Association
  for Computational Linguistics (Volume 1: Long Papers)}, pages 12073--12086,
  Bangkok, Thailand. Association for Computational Linguistics.

\end{thebibliography}

\clearpage

\onecolumn
\section*{Appendix}
\label{sec:appendix}

\appendix

\section{Cultural Context Example}
\label{appendix:AA_cultura_context_sample}
Table ~\ref{tab:cultural_context_annot_samples} shows the samples of cultural context annotations on KOLD dataset.

~\begin{table}[H]
  \resizebox{\columnwidth}{!}{%
    \begin{tabular}{@{}cclll@{}}
      \toprule
      \textbf{Category} &
      \textbf{Label} &
      \multicolumn{1}{c}{\textbf{Context(title)}} &
      \multicolumn{1}{c}{\textbf{Comment}} &
      \multicolumn{1}{c}{\textbf{Cultural Context Anntoation}} \\ \midrule
      \multirow{2}{*}{
        \begin{tabular}[c]{@{}l@{}}\textit{Cultural}\\ \textit{Knowledge}
      \end{tabular}} &
      OFF &
      \begin{tabular}[c]{@{}l@{}}\colorbox{blue!20}{GS25}, \colorbox{blue!20}{브레이브걸스} 포스터\\ 또 젠더 이슈 논란\\ \colorbox{blue!20}{GS25}, \colorbox{blue!20}{Brave Girls} poster, another\\ gender issue controversy
      \end{tabular} &
      \begin{tabular}[c]{@{}l@{}}\colorbox{blue!20}{브레이브걸스}=페미아이돌\\ \colorbox{blue!20}{Brave Girls}=feminist idol
      \end{tabular} &
      \begin{tabular}[c]{@{}l@{}}\textbf{GS25}: A major convenience store chain in South Korea.\\ They faced controversy over a promotional poster featuring\\ a hand reaching for a sausage, which some critics saw as\\ referencing a feminist symbol associated with Megalia.\\ \textbf{Brave Girls}: Known for their viral hit Rollin'.\\ Linked to feminist discourse in South Korea due to perceived\\ support for gender equality.
      \end{tabular} \\ \cmidrule(l){2-5}
      &
      NOT &
      \begin{tabular}[c]{@{}l@{}}`\colorbox{blue!20}{N번방}' 밝힌 `\colorbox{blue!20}{추적단불꽃}',\\ 與선대위 합류… 젠더 공약 돕는다\\ The `\colorbox{blue!20}{Tracking Team Fire}' that\\ exposed the `\colorbox{blue!20}{Nth Room}' joins the\\ ruling party's election committee...\\ Helping with gender pledges
      \end{tabular} &
      \begin{tabular}[c]{@{}l@{}}\colorbox{blue!20}{N번방}을 추적해 공론화시킨\\ \colorbox{blue!20}{추적단불꽃}을 응원합니다.\\ I support the \colorbox{blue!20}{tracking team Flame}\\ that tracked and publicized\\ the \colorbox{blue!20}{Nth Room}.
      \end{tabular} &
      \begin{tabular}[c]{@{}l@{}}\textbf{Nth Room}: A major digital sex crime scandal in South Korea\\ involving the production and distribution of explicit videos,\\ including those of minors, through Telegram chatrooms,\\ sparking nationwide outrage.\\ \textbf{Tracking Team Flame}: A duo of university student activists,\\ played a crucial role in exposing the case, leading to law\\ enforcement action and policy changes on digital sexual crimes.
      \end{tabular} \\ \midrule
      \multirow{2}{*}{
        \begin{tabular}[c]{@{}l@{}}\textit{Cultural}\\ \textit{Sentiment}
      \end{tabular}} &
      OFF &
      \begin{tabular}[c]{@{}l@{}}`모든 국가와 우호적 관계'…\\ 내부는 `공포정치'\\ `Friendly relations with all countries"\\ … internally `politics of fear'
      \end{tabular} &
      \begin{tabular}[c]{@{}l@{}}\colorbox{blue!20}{정은이가 친구}해 줄 거다.\\ \colorbox{blue!20}{Jeongeun will be your friend.}
      \end{tabular} &
      \begin{tabular}[c]{@{}l@{}}\textbf{`Jeongeun'} refers to North Korean leader Kim Jong-un.\\ He is known for ruling North Korea through a regime of\\ fear and political repression.
      \end{tabular} \\ \cmidrule(l){2-5}
      &
      NOT &
      \begin{tabular}[c]{@{}l@{}}박용진이 던진 `여성 군사훈련'…\\ 젠더 넘은 찬반 양론 불붙나\\ Yong-jin Park's proposal for\\ `women's military training'\\ ignites debate… Will the\\ controversy further intensify\\ beyond gender?
      \end{tabular} &
      \begin{tabular}[c]{@{}l@{}}\colorbox{blue!20}{남자도} 무조건 애낳으면\\ \colorbox{blue!20}{2년 육아 전담의무화}\\ \colorbox{blue!20}{Men should also be required to}\\ take \colorbox{blue!20}{full responsibility for childcare}\\ \colorbox{blue!20}{for two years} unconditionally\\ if they have a child.
      \end{tabular} &
      \begin{tabular}[c]{@{}l@{}}In South Korea's \textbf{gender debate}, men's two-year mandatory\\ military service is often compared to women’s role in childbirth\\ and the societal expectation of primary childcare responsibility.
      \end{tabular} \\ \midrule
      \multirow{2}{*}{
        \begin{tabular}[c]{@{}l@{}}\textit{Internet}\\ \textit{Culture}
      \end{tabular}} &
      OFF &
      \begin{tabular}[c]{@{}l@{}}{[}세상읽기{]} 여성+가족부 해체\\ {[}Reading the World{]} Dissolution\\ of the Ministry of Women\\ and Family
      \end{tabular} &
      \begin{tabular}[c]{@{}l@{}}응 준비완$\sim$$\sim$$\sim$\colorbox{blue!20}{군캉스} \colorbox{blue!20}{개꿀}$\sim$$\sim$$\sim$ \\ Yes, ready$\sim$$\sim$$\sim$ \\ \colorbox{blue!20}{Military vacation} \colorbox{blue!20}{so sweet}$\sim$$\sim$$\sim$
      \end{tabular} &
      \begin{tabular}[c]{@{}l@{}}\textbf{`Military vacation'}: a sarcastic term combining `military'\\ and `vacation,' used to criticize perceptions that South\\ Korea's mandatory military service is easier than it actually is.\\ \textbf{`So sweet'}: a slang term where `개' (dog) intensifies\\ `꿀' (sweet), meaning something is very easy or satisfying,\\ often used humorously or exaggeratedly.
      \end{tabular} \\ \cmidrule(l){2-5}
      &
      NOT &
      \begin{tabular}[c]{@{}l@{}}(재) 흑인 농부에게 쇠사슬에 묶여\\ 교육 당하는 중독자;\\ (Re) An addict being chained\\ and forced to receive education\\ by a Black farmer.
      \end{tabular} &
      \begin{tabular}[c]{@{}l@{}}\colorbox{blue!20}{두번째 댓글}\\ \colorbox{blue!20}{Second comment}
      \end{tabular} &
      \begin{tabular}[c]{@{}l@{}}\textbf{`Second comment'}: a common internet trend in South Korea\\ where users rush to comment early on articles or posts,\\ often just to claim a spot. It is typically meaningless and\\ unrelated to the original post.
      \end{tabular} \\ \bottomrule
    \end{tabular}%
  }
  \caption{Example of category labeling and cultural context annotations on KOLD. Label, Context(title), and Comment is from KOLD. (OFF: offensive, NOT: not offensive, \colorbox{blue!20}{blue}: culturally dependent content)}
  \label{tab:cultural_context_annot_samples}
\end{table}

\section{Pipeline Sample analysis}
\label{appendix:A_pipeline_sample_analysis}
Full Analysis on the samples used in pipeline in the following table. Chi-square analysis was done to prove / disprove the null hypothesis($H_0$): the accuracy of LLM-agree samples is independent of human agreement on both LLMs-Agree case and LLMs-Disagree case. The discussion on LLMs Agree sample was done in Section \ref{sec:quantitative_analysis_LLMerror}. For LLMs Agree samples(143 samples), the Chi-square value was 16.2064, and the p-value was 0.000057(<0.05). Therefore, for the samples that were resolved in step 2(LLMs Agree samples), the incorrectness is dependent with the difficulty of sample. For LLMs Disagree samples(28 samples), the Chi-square value was 0.0499, with the p-value 0.823218 (>0.05). Therefore, for samples that reached step 3, the null hypothesis($H_0$) failed to be rejected. \\

\begin{table}[h!]
  \centering
  \begin{minipage}[t]{0.5\textwidth}
    \centering
    \resizebox{\columnwidth}{!}{%
      \begin{tabular}{lcccc}
        \toprule
        & \multicolumn{2}{c}{LLMs Agree} & \multicolumn{2}{c}{LLMs Disagree}  \\
        Correct      & \cmark & \xmark  & \cmark  & \xmark       \\
        \midrule
        Human - Agree         & 91          & 14   & 10 &6  \\
        Human - Disagree         & 21 & 17   & 7          & 5 \\
        \bottomrule
      \end{tabular}
    }
    \caption{Analysis of 171 samples from the pipeline experiment. Samples were categorized based on whether the three LLMs in the second stage of the pipeline reached unanimous agreement. Each category was further divided by the Human majority voting outcome, assessing whether it aligned with the golden answers in the KOLD dataset. Finally, samples were analyzed based on agreement among the three Korean annotators of the KOLD dataset.}
    \label{table:pipeline_sample_analysis}
  \end{minipage}
\end{table}

\section{KOLD Annotators analysis}
\label{appendix:annotator_analysis}

\centering
\begin{minipage}[t]{0.7\textwidth}
  \centering
  \includegraphics[width=\linewidth]{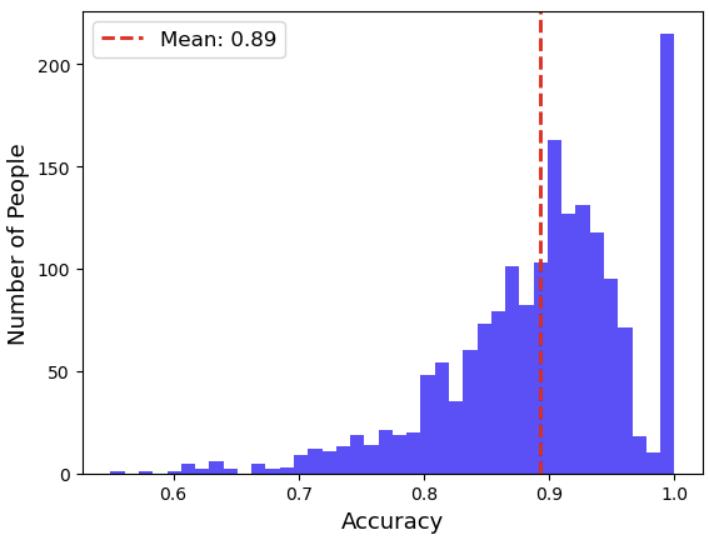}
  \captionof{figure}{Histogram showing the accuracy distribution of 1,749 Korean annotators(out of 3,124 Korean annotators, who annotated more than 9 samples) in the KOLD dataset.}
  \label{fig:kold_korean_stat}

\end{minipage}

\raggedright

\section{Prompt}
Appendix \ref{appendix:prompt_annotation} describes the prompts utilized to generated annotation. Appendix \ref{appendix:prompt_off_classify} is the prompt used for LLM to classify the offensiveness of the text.

\subsection{Prompt for Annotation}
\label{appendix:prompt_annotation}

This is the prompt selected for annotation generation. It consists of two steps: the RAG step and the annotation generation step. Together, this entire process represents a Chain-of-Thought (CoT) approach. \\

\subsubsection{RAG-step Prompt}
\label{appendix:prompt_rag}
Below is the original prompt used in \textbf{RAG step}, and prompt translated in English. The original prompt contains both English and Korean.
The parts that were originally written in English and hasn't been additionally translated are written in \textcolor{blue}{blue} font.\\
\vspace{10mm}

\textbf{Original Prompt}

\begin{tcolorbox}[colframe=gray, colback=gray!10, boxrule=0.5mm, sharp corners]
  다음 기사의 제목과 댓글에서 한국 문화나 한국의 인물, 또는 한국 인터넷 밈이 포함되어 있는 지 확인하고, 그 부분을 인터넷 검색 (web search) 후 관련된 한국 문화나 인물, 인터넷 밈에 대해서 인터넷 출처와 함께 알려줘.\\
  [1em]
  Title: \textbf{\{Article text title in Korean\}}\\
  Comment: \textbf{\{Article text comment in Korean\}}
\end{tcolorbox}

\textbf{Translated Prompt}

\begin{tcolorbox}[colframe=gray, colback=gray!10, boxrule=0.5mm, sharp corners]
  Check whether the following article's title and comments include any references to Korean culture, Korean figures, or Korean internet memes. Then, perform an internet search \textcolor{blue}{(web search)} on those parts and provide information about the related Korean culture, figures, or internet memes along with internet sources.\\[1em]
  \textcolor{blue}{Title:} \textbf{\{Article text title in Korean\}}\\
  \textcolor{blue}{Comment:} \textbf{\{Article text comment in Korean\}}
\end{tcolorbox}

\subsubsection{Annotation Generation-step Prompt}
\label{appendix:prompt_generation}
Below is the original prompt used in \textbf{annotation generation step}, and prompt translated in English. The original prompt contains both English and Korean. The parts that were originally written in English and hasn't been additionally translated are written in \textcolor{blue}{blue} font. The LLM must know the response of the previous prompt(\textbf{RAG step} prompt) before given the following prompt.\\
\vspace{5mm}

\textbf{Original Prompt}

\begin{tcolorbox}[colframe=gray, colback=gray!10, boxrule=0.5mm, sharp corners]
  외국인이 이해할 수 있도록 예시처럼 영어로 간결하게 설명해줘.

예시)

Title: ""Suicide bombing at a mosque in northwest Pakistan... 'Over 40 dead'""

Comment: If Dori Dori gets elected, you'll end up like that.

- ""Dori Dori"": is a nickname used in South Korea to refer to President Yoon Suk-yeol. This nickname originated from his habit of moving his head side to side during speeches or public appearances. The term became an internet meme, often used in a political context to mock or criticize him.

설명은 객관적이어야만 해. 아래의 번역된 기사 제목과 댓글을 참고하도록 해.\\
Title: \textbf{\{Article text title in translated in English\}}\\
Comment: \textbf{\{Article text comment translated in English\}}
\end{tcolorbox}

\pagebreak

\textbf{Translated Prompt}

\begin{tcolorbox}[colframe=gray, colback=gray!10, boxrule=0.5mm, sharp corners]
Explain it concisely in English so that foreigners can understand, like in the example below.

Example)

{\color{blue}
Title: ""Suicide bombing at a mosque in northwest Pakistan... 'Over 40 dead'""

Comment: If Dori Dori gets elected, you'll end up like that.

- ""Dori Dori"": is a nickname used in South Korea to refer to President Yoon Suk-yeol. This nickname originated from his habit of moving his head side to side during speeches or public appearances. The term became an internet meme, often used in a political context to mock or criticize him.}

The explanation must remain objective. Refer to the translated article title and comments below.

\textcolor{blue}{Title:} \textbf{\{Article text title translated in English\}}

\textcolor{blue}{Comment:} \textbf{\{Article text comment translated in English\}}
\end{tcolorbox}

\subsection{Offensiveness Classification Prompt}
\label{appendix:prompt_off_classify}

This prompt is used to guide an LLM in evaluating the offensiveness of a given text by utilizing external information (annotations generated by the prompt in Appendix \ref{appendix:prompt_annotation}). This is also the prompt used in the second stage of the pipeline.

\begin{tcolorbox}[colframe=gray, colback=gray!10, boxrule=0.5mm, sharp corners, breakable]
  Consider the following definitions.\\
  1. offensive: the text which contains one of the following type:  \\
  (a) Word Modification: Instances where offensive language is altered to evade censorship but remains clearly recognizable as an insult.\\
  (b) Reverse Discrimination Statements: Remarks expressing dissatisfaction with privileges or benefits attributed to a particular group, while portraying the speaker's group as a victim of discrimination.\\
  (c) Incitement of Violence: Statements that incite or support violence or hate groups.\\
  (d) Derogatory Remarks: Statements that imply or state that a particular group or individual is inferior to others. This includes dehumanizing language, such as comparing people to animals, insects, objects, etc.\\
  (e) Satire: Statements that use sarcasm, humor, or satire to attack a particular group or individual.\\
  (f) Stereotyping / Fact Distortion: Statements associating a specific group with negative attributes, like murder or terrorism, or delivering unsubstantiated and negatively distorted information about a specific group.\\
  2. normal: the text which is not offensive and adheres to social norms.

  Classify the post delimited by two backticks as offensive and normal. It is a Korean-written corpus translated in English. Refer to the annotation in your decision.\\
  If it is offensive, span the parts(sentence, phrase or word) which you think it is offensive.

  Example output for offensive : \\
  Offensiveness : True\\
  Span : [span]

  Example output for normal : \\
  Offensiveness : False

  `Title: \{\textbf{Article text title translated in English}\} \\
  Comment: \{\textbf{Article text comment translated in English}\} `

  Annotation: \\
  \{\textbf{Annotation generated by the prompt in Appendix \ref{appendix:prompt_annotation}}\}
\end{tcolorbox}

\pagebreak

\section{Annotation Generation Example}
\label{appendix:D_annotation_generation_example}
The following is an example of the generated annotation. The given title and comments are as follows:
\begin{tcolorbox}[colframe=gray, colback=gray!10, boxrule=0.5mm, sharp corners, breakable]
  \textbf{Title (KOR)}: [SNS세상] ""노래할 땐 사투리 안 쓰냐고?…지방말투 차별 마세요""\\
  \textbf{Comment (KOR)}: 무까끼하이

  \textbf{Title (ENG)}: [SNS World] ""Don't you speak dialect when you sing?... Don't discriminate against regional accents""\\
  \textbf{Comment (ENG)}: Mukakki high
\end{tcolorbox}

\subsection{Response of RAG-step prompt}
This is the response of the LLM after the \textbf{RAG-step} Prompt in Appendix \ref{appendix:prompt_rag}.
Since the original prompt is written mostly in Korean, and the given input title and comment is in Korean, the output response of the LLM is also in Korean. Therefore, the translated response is also given below. The parts that were originally written in English and hasn't been additionally translated are written in \textcolor{blue}{blue} font. \\

\vspace{5mm}

\textbf{Original Response}

\begin{tcolorbox}[colframe=gray, colback=gray!10, boxrule=0.5mm, sharp corners, breakable]
  제목과 댓글에서 한국 문화와 관련된 요소를 확인한 결과, 댓글에 '무까끼하이'라는 표현이 등장합니다. '무까끼하이'는 경상도 사투리로 '무식하게'라는 의미를 지니며, 힙합 듀오 메타와 렉스(Meta \& Wreckx)의 곡 제목이기도 합니다. 이 곡은 음악 산업에서 돈만을 추구하는 이들을 비판하는 내용을 담고 있으며, 전체 가사가 경상도 사투리로 구성되어 있습니다. (출처: 마니아DB)

  또한, 이 곡은 2012년 한국대중음악상에서 최우수 랩\&힙합 노래 부문을 수상하며 그 가치를 인정받았습니다. (출처: KOREAN MUSIC AWARDS)

  이처럼 '무까끼하이'는 한국 힙합 문화에서 사투리의 독특한 표현을 활용한 사례로, 음악 산업에 대한 비판적 시각을 담고 있어 한국 대중음악사에서 의미 있는 작품으로 평가받고 있습니다.
\end{tcolorbox}

\textbf{Translated Response}
\begin{tcolorbox}[colframe=gray, colback=gray!10, boxrule=0.5mm, sharp corners, breakable]
  As a result of checking the title and comments for elements related to Korean culture, the comment includes the phrase 'Mukakki High.' 'Mukakki High' is a phrase from the Gyeongsang-do dialect meaning 'recklessly' or 'ignorantly.' It is also the title of a song by the hip-hop duo Meta and Wreckx\textcolor{blue}{(Meta \& Wreckx)}. The song criticizes those in the music industry who pursue money above all else, and its entire lyrics are written in the Gyeongsang-do dialect. (Source: ManiaDB)

  Furthermore, this song was recognized for its value by winning the Best Rap \& Hip-Hop Song award at the 2012 Korean Music Awards. (Source: \textcolor{blue}{Korean Music Awards})

  In this way, 'Mukakki High' is an example of utilizing unique expressions from regional dialects in Korean hip-hop culture. It conveys a critical perspective on the music industry and is regarded as a significant work in the history of Korean popular music.
\end{tcolorbox}

\subsection{Response of annotation generation-step prompt}
This is the response of the LLM after \textbf{the annotation generation step} prompt in Appendix \ref{appendix:prompt_generation}.
This prompt asks to generate annotation in English, therefore the original response is in English.

\begin{tcolorbox}[colframe=gray, colback=gray!10, boxrule=0.5mm, sharp corners, breakable]
  Title: [SNS World] "Don't you speak dialect when you sing?... Don't discriminate against regional accents"\\
  Comment: Mukakki high

  \textbf{"Mukakki high"}: This phrase originates from a regional dialect in South Korea, specifically the Gyeongsang Province dialect, meaning "recklessly" or "brashly." It gained popularity through a song by the South Korean hip-hop duo Meta \& Wreckx titled "Mukakki High," which uses Gyeongsang dialect throughout its lyrics. The song criticizes the greed-driven nature of the music industry and became a cultural reference in Korean hip-hop.
\end{tcolorbox}

\section{LLM Offensiveness Classification Example}
\label{appendix:E_off_class_example}
The following is an example of the classification response. The given title, comments, generated annotations, and the golden answer are as follows: \\
\begin{tcolorbox}[colframe=gray, colback=gray!10, boxrule=0.5mm, sharp corners, breakable]
  \textbf{Title (KOR)}: [자막뉴스] '노마스크' 백인 남성에 "마트에서 나가라" 본때 보여준 한인들 / JTBC News\\
  \textbf{Comment (KOR)}: 하다하다 이런걸로 국뽕질이네\\
  \textbf{Title (ENG)}: [Subtitled News] Koreans Teach a Lesson to "No Mask" White Man with "Get Out of the Store" / JTBC News\\
  \textbf{Comment (ENG)}: Doing this kind of thing out of \textbf{extreme national pride}, seriously.\\
  \textbf{Annotation (ENG)}:
  \begin{quote}
    \textbf{"Extreme national pride (국뽕)"}:
    This term is a Korean internet slang combining "nation" and "methamphetamine," used to mock or criticize excessive patriotism or over-the-top national pride.
    It implies being overly proud of one's country to the point of irrationality or exaggeration.
    The term is popular in online discussions in South Korea, often appearing in contexts where people feel certain actions or statements are driven by undue nationalistic sentiment.
  \end{quote}
  \textbf{Offensive?} : TRUE
\end{tcolorbox}

This is the response provided by the LLM when tasked with classifying the offensiveness of a post. The classification was conducted using external information (generated annotations), as outlined in the prompt included in Appendix \ref{appendix:prompt_off_classify}.
\begin{tcolorbox}[colframe=gray, colback=gray!10, boxrule=0.5mm, sharp corners, breakable]
  Offensiveness : True\\
  Span : ["Doing this kind of thing out of extreme national pride"]
\end{tcolorbox}

\end{document}